\documentclass[10pt,twocolumn,letterpaper]{article}

\pdfoutput=1

\usepackage{cvpr}
\usepackage{times}
\usepackage{epsfig}
\usepackage{graphicx}
\usepackage{amsmath}
\usepackage{amssymb}
\usepackage{adjustbox}
\usepackage{arydshln}
\usepackage[utf8]{inputenc}
\usepackage{caption}
\usepackage{subcaption}

\usepackage[numbers,sort]{natbib}

\newif\ifkeepComments
\keepCommentstrue

\newif\ifArXivVersion
\ArXivVersiontrue

\newcommand{\isArXiv}[2]{\ifArXivVersion #1\else #2 \fi}

\renewcommand{\paragraph}[1]{\smallskip\noindent{\bf{#1}}}

\graphicspath{{images/}{images_reb/}{images_sup/}}

\isArXiv{\usepackage[pagebackref=true,breaklinks=true,letterpaper=true,colorlinks,bookmarks=false]{hyperref} 
\newif\ifbackrefshowonlyfirst
\backrefshowonlyfirsttrue
\makeatletter
\let\BR@direct@old@hyper@natlinkstart\hyper@natlinkstart
\renewcommand*{\hyper@natlinkstart}{\phantomsection\BR@direct@old@hyper@natlinkstart}
\let\BR@direct@oldBR@citex\BR@citex
\renewcommand*{\BR@citex}{\phantomsection\BR@direct@oldBR@citex}%

\long\def\hyper@page@BR@direct@ref#1#2#3{\hyperlink{#3}{#1}}

\ifx\backrefxxx\hyper@page@backref
    \let\backrefxxx\hyper@page@BR@direct@ref
    \ifbackrefshowonlyfirst
    \fi
\else
    \ifbackrefshowonlyfirst
    \fi
\fi

\RequirePackage{etoolbox}
\patchcmd{\Hy@backout}{Doc-Start}{\@currentHref}{}{\errmessage{I can't seem to patch backref}}
\makeatother
}{\usepackage[breaklinks=true,bookmarks=false]{hyperref}}

\hypersetup{pdfauthor={Ignacio Rocco},pdftitle={Convolutional neural network architecture for geometric matching}}

\makeatletter
\renewcommand*{\@fnsymbol}[1]{\ensuremath{\ifcase#1\or \or \or \or
    \or \or \or \or \or \else\@ctrerr\fi}}
\makeatother

\cvprfinalcopy 

\def\cvprPaperID{2704} 

\isArXiv{\setcounter{page}{1}}{\ifcvprfinal\pagestyle{empty}\fi}

\begin{document}

\title{Convolutional neural network architecture for geometric matching}

\author{Ignacio Rocco$^{1,2}$ \qquad Relja Arandjelovi\'{c}\,$^{1,2,*}$
 \qquad Josef Sivic$^{1,2,3}$\\
$^1$DI ENS \quad \quad \quad \quad $^2$INRIA\quad \quad \quad \quad $^3$CIIRC
\thanks{$^1$Département d’informatique de l’ENS, École normale supérieure, CNRS, PSL Research University, 75005 Paris, France.}
\thanks{$^3$Czech Institute of Informatics, Robotics and Cybernetics at the Czech Technical University in Prague.}
\thanks{$^*$Now at DeepMind.}
}

\maketitle
\isArXiv{}{\thispagestyle{empty}}

\begin{abstract}
We address the problem of determining correspondences between two images in agreement with a geometric model such as an affine or thin-plate spline transformation, and estimating its parameters. The contributions of this work are three-fold. 
First, we propose a convolutional neural network architecture for geometric matching.   The architecture is based on three main components that mimic the standard steps of feature extraction, matching and simultaneous inlier detection and model parameter estimation, while being trainable end-to-end. 
Second, we demonstrate that the network parameters can be trained from synthetically generated imagery without the need for manual annotation and that our matching layer significantly increases generalization capabilities to never seen before images. Finally, we show that the same model can perform both instance-level and category-level matching giving state-of-the-art results on the challenging Proposal Flow dataset.
\end{abstract}

\section{Introduction}

Estimating correspondences between images is one of the fundamental problems in computer vision~\cite{forsyth2002computer,hartley2003multiple} with applications ranging from large-scale 3D reconstruction~\cite{agarwal2009building} to image manipulation \cite{HaCohen11} and semantic segmentation \cite{rubinstein2013unsupervised}. Traditionally, correspondences consistent with a geometric model such as epipolar geometry or planar affine transformation, are computed by detecting and matching local features (such as SIFT~\cite{lowe2004distinctive} or HOG~\cite{Dalal05,ham2016}), followed by pruning incorrect matches using local geometric constraints~\cite{schmid1997local,Sivic03} and robust estimation of a global geometric transformation using algorithms such as RANSAC \cite{Fischler81} or Hough transform~\cite{Lamdan88,Leibe08,lowe2004distinctive}. This approach works well in many cases but fails in situations that  exhibit (i) large changes of depicted appearance due to \eg\ intra-class variation~\cite{ham2016}, or (ii) large changes of scene layout or non-rigid deformations that require complex geometric models with many parameters
which are hard to estimate in a manner robust to outliers.

In this work we build on the traditional approach and develop a convolutional neural network (CNN) architecture that mimics the standard matching process. First, we replace the standard local features with powerful trainable convolutional neural network features~\cite{Krizhevsky12,Simonyan15}, which allows us to handle large changes of appearance between the matched images. Second, we develop trainable matching and transformation estimation layers that can cope with noisy and incorrect matches in a robust way, mimicking the good practices in feature matching such as the second nearest neighbor test \cite{lowe2004distinctive}, neighborhood consensus \cite{schmid1997local,Sivic03} and Hough transform-like estimation \cite{Lamdan88,Leibe08,lowe2004distinctive}. 

The outcome is a convolutional neural network architecture trainable for the end task of geometric matching, which can handle large appearance changes, and is therefore suitable for both instance-level and category-level matching problems.

\begin{figure}
    \centering
\centering \includegraphics[width=\columnwidth]{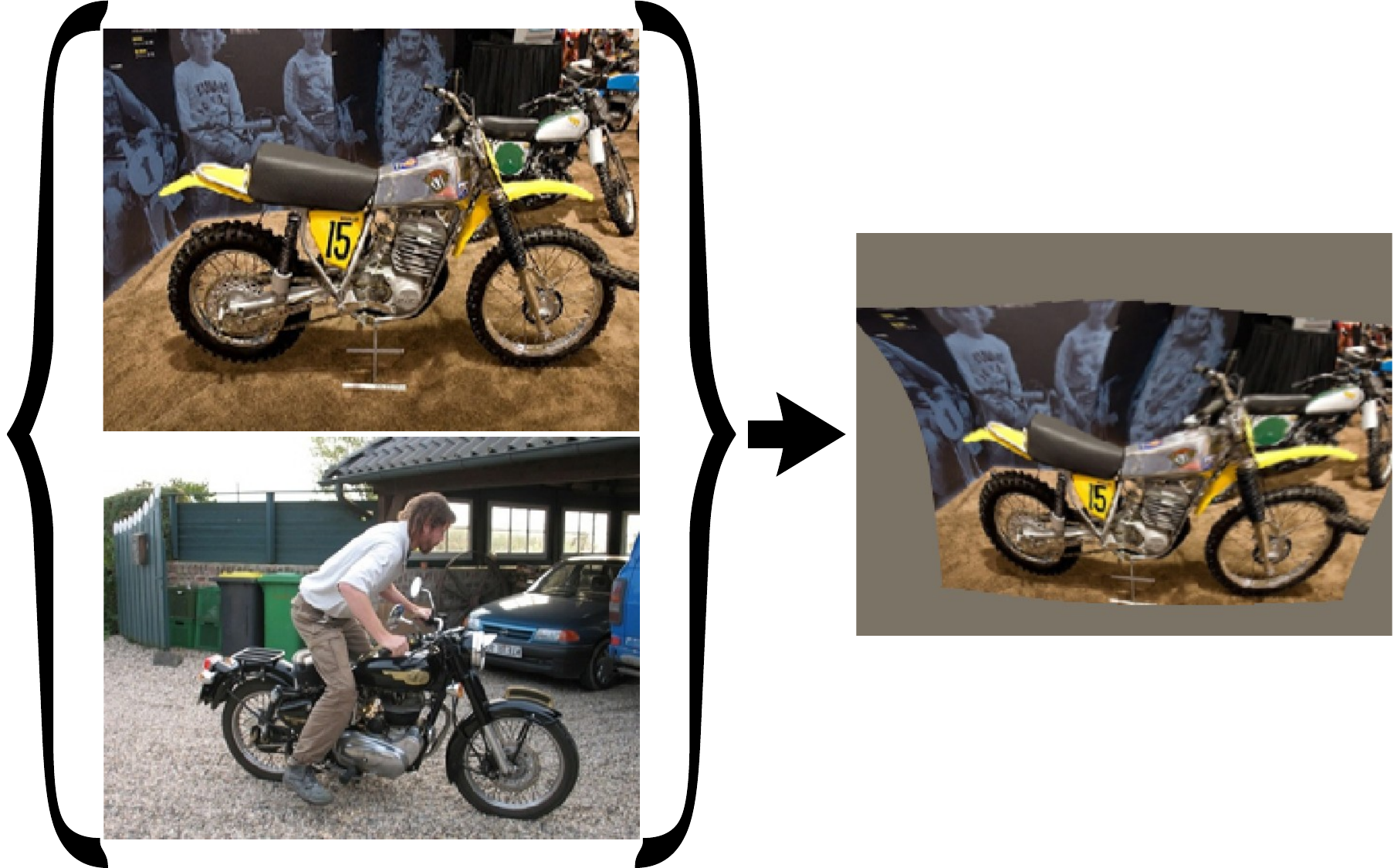}
\captionsetup{font={small}}
\vspace*{-6mm}\caption{Our trained geometry estimation network automatically aligns two images with substantial appearance differences.
It is able to estimate large deformable transformations robustly in the presence of clutter.}
    \label{fig:teaser}
    \vspace{-0.4cm}
\end{figure}

\section{Related work}

The classical approach for finding correspondences involves identifying interest points and computing local descriptors around these points \cite{harris1988combined,schmid1997local,lowe1999object,mikolajczyk2002affine,lowe2004distinctive,Berg05,bay2006surf}. 
While this approach performs relatively well for instance-level matching, the feature detectors and descriptors lack the generalization ability for category-level matching.

Recently, convolutional neural networks have been used to learn powerful feature descriptors which are more robust to appearance changes than the classical descriptors \cite{jahrer2008learned,simo2015discriminative,han2015matchnet,zagoruyko2015learning,balntas2016pn}.
However, these works still divide the image into a set of local patches and 
extract a descriptor individually from each patch. 
Extracted descriptors are then compared with an appropriate distance measure \cite{jahrer2008learned,simo2015discriminative,balntas2016pn}, by directly outputting a similarity score \cite{han2015matchnet,zagoruyko2015learning}, or even by directly outputting a binary \emph{matching/non-matching} decision \cite{altwaijry2016learning}.

In this work, we take a different approach, treating the image as a whole, instead of a set of patches. Our approach has the advantage of capturing the interaction of the different parts of the image in a greater extent, which is not possible when the image is divided into 
a set of local regions.

Related are also network architectures for estimating inter-frame motion in video~\cite{weinzaepfel2013deepflow,fischer2015flownet,thewlis16fully-trainable} or instance-level homography estimation~\cite{detone2016deep}, however their goal is very different from ours, targeting high-precision correspondence with very limited appearance variation and background clutter. Closer to us is the network architecture of~\cite{kanazawa2016warpnet} which, however, tackles a different problem of fine-grained category-level matching (different species of birds) with limited background clutter and small translations and scale changes, as their objects are largely centered in the image.
In addition, their architecture is based on a different matching layer, which we show not to perform as well as the matching layer used in our work.

Some works, such as \cite{Berg05,liu2011sift,Duchenne11,Kim13,long2014convnets,ham2016}, have
addressed the hard problem of category-level matching, but rely on traditional non-trainable optimization for matching \cite{Berg05,liu2011sift,Duchenne11,Kim13,long2014convnets}, or
guide the matching using object proposals \cite{ham2016}.
On the contrary, our approach is fully trainable in an end-to-end manner and does not require
any optimization procedure at evaluation time, or guidance by object proposals.

Others \cite{learned-miller2006,shokrollahi2015unsupervised,Zhou15} have addressed the problems of instance and category-level correspondence by performing joint image alignment. However, these methods differ from ours as they:
(i) require class labels;
(ii) don't use CNN features; 
(iii) jointly align a large set of images, while we align image pairs;
and (iv) don't use a trainable CNN architecture for alignment as we do.

\section{Architecture for geometric matching}

In this section, we introduce a new convolutional neural network 
architecture for estimating parameters of a geometric transformation between two input images.
The architecture is designed to mimic the classical computer vision pipeline (\eg\ \cite{philbin2007object}), while using differentiable modules so that it is trainable end-to-end for the geometry estimation task.
The classical approach consists of the following stages:
(i) local descriptors (\eg\ SIFT) are extracted from both input images,
(ii) the descriptors are matched across images to form a set of tentative correspondences,
which are then used to
(iii) robustly estimate the parameters of the geometric model using RANSAC or Hough voting.

\begin{figure}
\centering \includegraphics[width=\columnwidth]{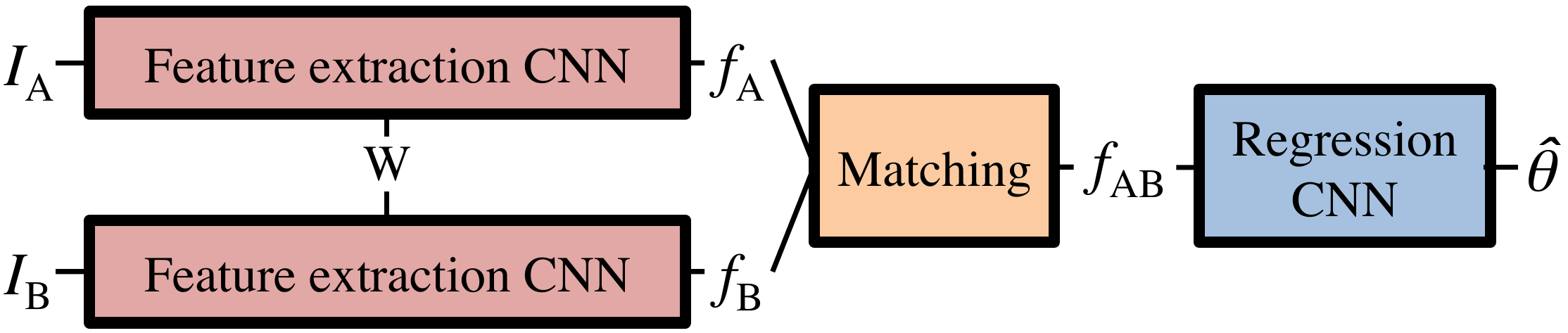} 
\captionsetup{font={small}}
\caption{{\bf Diagram of the proposed architecture.}
Images $I_A$ and $I_B$ are passed through feature extraction networks which have
tied parameters $W$, followed by a matching network which matches the descriptors.
The output of the matching network is passed through a regression network which outputs
the parameters of the geometric transformation.} \label{fig:proposed-arch}
\vspace{-0.4cm}
\end{figure}

Our architecture, illustrated in Fig.\,\ref{fig:proposed-arch}, mimics this process by:
(i) passing input images $I_A$ and $I_B$ through a siamese architecture consisting of convolutional layers, thus extracting feature maps $f_A$ and $f_B$ which are analogous to dense local descriptors,
(ii) matching the feature maps (``descriptors'') across images into a tentative correspondence map $f_{AB}$, followed by a
(iii) regression network which directly
outputs the parameters of the geometric model, $\hat{\theta}$, in a robust manner.
The inputs to the network are the two images, and the outputs are the parameters
of the chosen geometric model, \eg\ a 6-D vector for an affine transformation.

In the following, we describe each of the three stages in detail.

\subsection{Feature extraction}

The first stage of the pipeline is feature extraction, for which we use a standard CNN architecture.
A CNN without fully connected layers takes an input image and produces
a feature map $f \in \mathbb{R}^{h \times w \times d}$, which can be
interpreted as a $h \times w$ dense spatial grid of $d$-dimensional local descriptors.
A similar interpretation has been used previously in instance retrieval~\cite{Azizpour14,Babenko15,Gong14,arandjelovic2015netvlad} demonstrating high discriminative
power of CNN-based descriptors.
Thus, for feature extraction we use the VGG-16 network \cite{Simonyan15}, cropped at the \texttt{pool4} layer (before the ReLU unit),
followed by per-feature L2-normalization. We use a pre-trained model, originally trained on ImageNet \cite{deng2009imagenet} for the task of image classification.
As shown in Fig.\,\ref{fig:proposed-arch}, the feature extraction network is duplicated and arranged in a
siamese configuration such that the two input images are passed through two identical networks which
share parameters.

\subsection{Matching network}
\label{sec:fusion}

The image features produced by the feature extraction networks should be combined into a single tensor
as input to the regressor network to estimate the geometric transformation.
We first describe the classical approach for generating tentative correspondences, and then
present our matching layer which mimics this process.

\paragraph{Tentative matches in classical geometry estimation.}
Classical methods start by
computing similarities between all pairs of descriptors across the two images.
From this point on, the original descriptors are discarded as all the necessary information for
geometry estimation is contained in the pairwise descriptor similarities and their spatial locations.
Secondly, the pairs are pruned by either thresholding the similarity values, or, more commonly,
only keeping the matches which involve the nearest (most similar) neighbors.
Furthermore, the second nearest neighbor test \cite{lowe2004distinctive} prunes the matches
further by requiring that the match strength is significantly stronger
than the second best match involving the same descriptor, which is very effective at discarding
ambiguous matches. 

\paragraph{Matching layer.}
Our matching layer applies a similar procedure.
Analogously to the classical approach, only descriptor similarities
and their spatial locations should be considered for geometry estimation, and not the original descriptors themselves.

To achieve this, we propose to use a \emph{correlation layer} followed by normalization. Firstly, all pairs of similarities between descriptors are computed in the correlation layer.
Secondly, similarity scores are processed and normalized such that ambiguous matches are strongly down-weighted.

\begin{figure}
\centering \includegraphics[width=0.95\columnwidth]{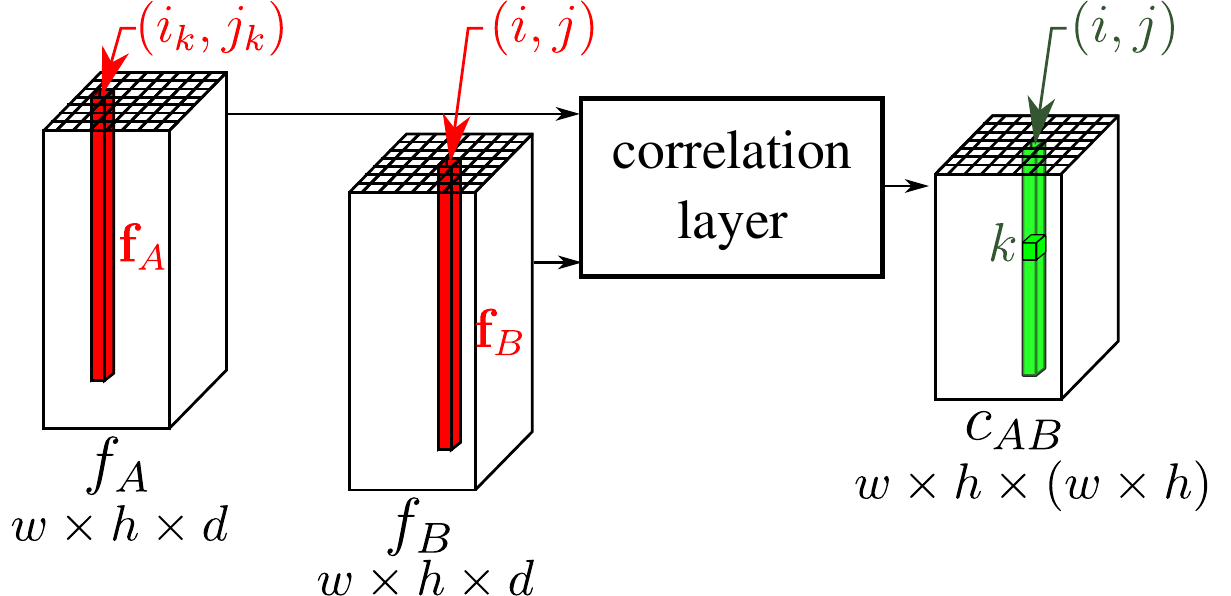} 
\captionsetup{font={small}}
\vspace{-0.2cm}
\caption{{\bf Correlation map computation with CNN features.}
The correlation map $c_{AB}$ contains all pairwise similarities between individual features $\mathbf{f}_A\in f_A$ and $\mathbf{f}_B\in f_B$. At a particular spatial location $(i,j)$ the correlation map output $c_{AB}$ contains all the similarities between $\mathbf{f}_B(i,j)$ and all $\mathbf{f}_A\in f_A$.
}
\label{fig:correlation}
\vspace{-0.25cm}
\end{figure}

In more detail, given L2-normalized dense feature maps $f_A,f_B\in\mathbb{R}^{h\times w
\times d}$, the correlation map $c_{AB} \in \mathbb{R}^{h\times w \times (h\times w)}$ outputted by the correlation layer contains at each position the scalar product of a pair of individual descriptors $\mathbf{f}_A\in f_A$ and $\mathbf{f}_B\in f_B$, as detailed in Eq.~\eqref{eq:correlation}.

\begin{equation}
\begin{aligned}
c_{AB}(i,j,k) = \mathbf{f}_B(i,j)^T\mathbf{f}_A(i_k,j_k)
\end{aligned}
\label{eq:correlation}
\end{equation}

\noindent where $(i,j)$ and $(i_k,j_k)$ indicate the individual feature positions in the $h\times w$ dense feature maps, and $k=h(j_k-1)+i_k$ is an auxiliary indexing variable for $(i_k,j_k)$.

A diagram of the correlation layer is presented in Fig.\,\ref{fig:correlation}. Note that at a particular position $(i,j)$, the correlation map $c_{AB}$ contains the similarities between $\mathbf{f}_B$ at that position and \emph{all} the features of $f_A$.

As is done in the classical methods for tentative correspondence estimation, it is important to postprocess
the pairwise similarity scores to remove ambiguous matches. To this end, we apply a channel-wise normalization
of the correlation map at each spatial location to produce the final tentative correspondence map $f_{AB}$.
The normalization is performed by ReLU, to zero out negative correlations, followed by L2-normalization, which has two desirable effects.
First, let us consider the case when descriptor $\mathbf{f}_B$ correlates well with only a single feature in $f_A$.
In this case, the normalization will amplify the score of the match, akin to the nearest neighbor matching in
classical geometry estimation.
Second, in the case of the descriptor $\mathbf{f}_B$ matching multiple features in $f_A$ due to the existence
of clutter or repetitive patterns, matching scores will be down-weighted similarly to the second nearest neighbor
test \cite{lowe2004distinctive}. However, note that both the correlation and the normalization operations are differentiable with respect
to the input descriptors, which facilitates backpropagation thus enabling end-to-end learning.

\paragraph{Discussion.}
The first step of our matching layer, namely the correlation layer, is somewhat similar to layers used in DeepMatching \cite{weinzaepfel2013deepflow} and FlowNet \cite{fischer2015flownet}.
However, DeepMatching \cite{weinzaepfel2013deepflow} only uses deep  RGB patches
and no part of their architecture is trainable. FlowNet \cite{fischer2015flownet} uses a spatially constrained
correlation layer such that similarities are are only computed in a restricted spatial neighborhood
thus limiting the range of geometric transformations that can be captured.
This is acceptable for their task of learning to estimate optical flow, but is inappropriate for larger transformations that we consider in this work.
Furthermore, neither of these methods performs score normalization, which we find to be crucial in dealing with cluttered scenes.

Previous works have used other matching layers to combine descriptors across images, namely
simple concatenation of descriptors along the channel dimension \cite{detone2016deep} or subtraction \cite{kanazawa2016warpnet}.
However, these approaches suffer from two problems. First, as following layers are typically convolutional,
these methods also struggle to handle large transformations as they are unable to detect
long-range matches.
Second, when concatenating or subtracting descriptors, instead of computing pairwise descriptor similarities as is commonly done in classical geometry estimation and mimicked by the correlation layer, image content information is directly outputted.
To further illustrate why this can be problematic, consider two pairs of images that are related with the same geometric transformation -- the concatenation
and subtraction strategies will produce different outputs for the two cases, making it hard for the regressor
to deduce the geometric transformation. In contrast, the correlation layer output is likely to produce
similar correlation maps for the two cases, regardless of the image content, thus simplifying the problem
for the regressor.
In line with this intuition, in Sec.\,\ref{sec:generalization} we show that the concatenation and subtraction
methods indeed have difficulties generalizing beyond the training set,
while our correlation layer achieves generalization yielding superior results.

\subsection{Regression network}
The normalized correlation map is passed through a regression network which directly estimates parameters
of the geometric transformation relating the two input images.
In classical geometry estimation, this step consists of robustly estimating the transformation from the
list of tentative correspondences. Local geometric constraints are often used to further prune the list
of tentative matches \cite{schmid1997local,Sivic03} by only retaining matches which are consistent
with other matches in their spatial neighborhood.
Final geometry estimation is done by
RANSAC \cite{Fischler81} or Hough voting \cite{Lamdan88,Leibe08,lowe2004distinctive}.

\begin{figure}
    \centering
  \includegraphics[width=\columnwidth]{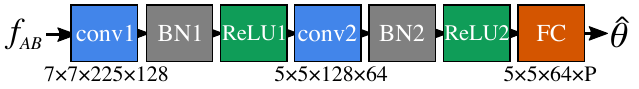} 
  \captionsetup{font={small}}
  \vspace{-0.6cm}
    \caption{{\bf Architecture of the regression network.} It is composed of two convolutional layers without padding and stride equal to 1, followed by batch normalization and ReLU, and a final fully connected layer which regresses to the $P$ transformation parameters.}
    \label{fig:reg-nets}
    \vspace*{-0.3cm}
\end{figure}

We again mimic the classical approach using a neural network,
where we stack two blocks of convolutional layers,
followed by batch normalization~\cite{ioffe2015batch} and the ReLU non-linearity, and add a final fully connected
layer which regresses to the parameters of the transformation, as shown in Fig.\,\ref{fig:reg-nets}.
The intuition behind this architecture is that the estimation is performed in a bottom-up manner
somewhat like Hough voting, where early convolutional layers vote for candidate transformations,
and these are then processed by the later layers to aggregate the votes.
The first convolutional layers can also enforce local neighborhood consensus \cite{schmid1997local,Sivic03}
by learning filters which only fire if nearby descriptors in image A are matched to nearby descriptors in image B,
and we show qualitative evidence in Sec.\,\ref{sec:whatislearned} that this indeed does happen.

\paragraph{Discussion.}
A potential alternative to a convolutional regression network is to use fully connected layers.
However, as the input correlation map size is quadratic in the number of image features, such a network
would be hard to train due to a large number of parameters that would need to be learned,
and it would not be scalable due to occupying too much memory and being too slow to use.
It should be noted that even though the layers in our architecture are convolutional,
the regressor can learn to estimate large transformations.
This is because one spatial location in the correlation map contains similarity scores between the corresponding
feature in image B and \emph{all} the features in image A (c.f.\ equation \eqref{eq:correlation}),
and not just the local neighborhood as in \cite{fischer2015flownet}.

\subsection{Hierarchy of transformations}
Another commonly used approach when estimating image to image transformations is to start by estimating a simple
transformation and then progressively increase the model complexity, refining the estimates along the way \cite{lowe1999object,Berg05,philbin2007object}.
The motivation behind this method is that estimating a very complex transformation could be hard
and computationally inefficient in the presence of clutter, so a robust and fast rough estimate of
a simpler transformation can be used as a starting point, also regularizing the subsequent
estimation of the more complex transformation.

\begin{figure*}
    \centering
\includegraphics[width=\linewidth]{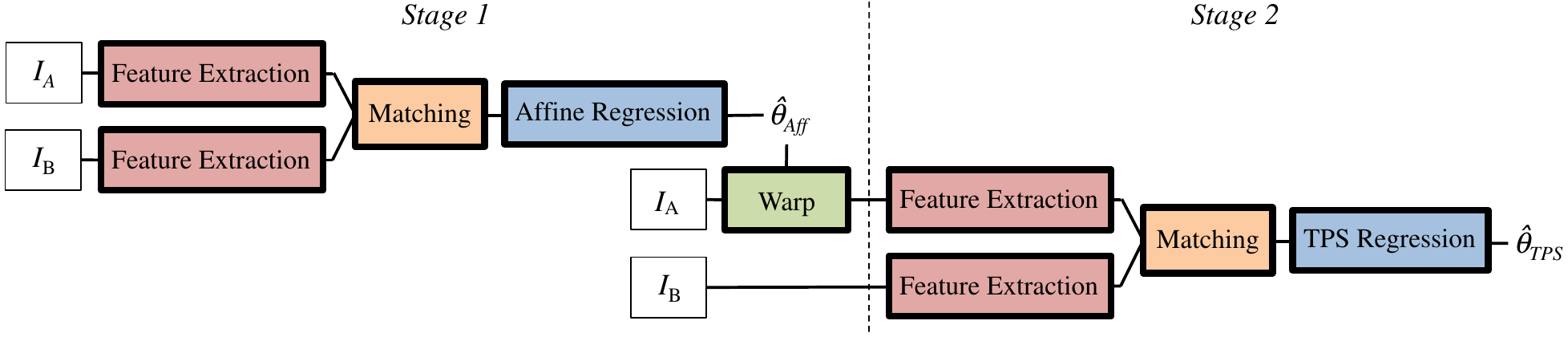} 
\captionsetup{font={small}}
\vspace*{-0.65cm}
    \caption{{\bf Estimating progressively more complex geometric transformations.}
Images A and B are passed through a network which estimates an affine transformation with parameters $\hat{\theta}_{\text{Aff}}$ (see Fig.\,\ref{fig:proposed-arch}).
Image A is then warped using this transformation to roughly align with B,
and passed along with B through a second network
which estimates a thin-plate spline (TPS) transformation that refines the alignment.
}
    \label{fig:hierarchy}
    \vspace{-0.3cm}
\end{figure*}

We follow the same good practice and start by estimating an affine transformation, which is a 6 degree of freedom
linear transformation capable of modeling translation, rotation, non-isotropic scaling and shear.
The estimated affine transformation is then used to align image B to image A using
an image resampling layer \cite{Jaderberg15}. The aligned images are then passed
through a second geometry estimation network which estimates 18 parameters of
a thin-plate spline transformation. The final estimate of the geometric transformation
is then obtained by composing the two transformations, which is also a thin-plate spline.
The process is illustrated in Fig.\,\ref{fig:hierarchy}.

\section{Training}

In order to train the parameters of our geometric matching CNN,
it is necessary to design the appropriate loss function, and to use suitable training data.
We address these two important points next.

\subsection{Loss function}

We assume a fully supervised setting, where the training data consists of pairs of images and the desired outputs
in the form of the parameters $\theta_{GT}$ of the ground-truth geometric transformation.
The loss function $\mathcal{L}$ is designed to compare the estimated transformation $\hat{\theta}$ with the ground-truth transformation $\theta_{GT}$ and, more importantly, compute the gradient of the loss function with respect to the estimates $\frac{\partial \mathcal{L}}{\partial \hat{\theta}}$.
This gradient is then used in a standard manner to learn the network parameters which minimize the loss function
by using backpropagation and Stochastic Gradient Descent.

It is desired for the loss to be general and not specific to a particular type of geometric model, so that
it can be used for estimating affine, homography, thin-plate spline or any other geometric transformation.
Furthermore, the loss should be independent of the parametrization of the transformation and thus should not
directly operate on the parameter values themselves.
We address all these design constraints by measuring loss on an imaginary grid of points which is being deformed
by the transformation. Namely, we construct a grid of points in image A, transform it using the ground truth
and neural network estimated transformations $\mathcal{T}_{\theta_{GT}}$ and $\mathcal{T}_{\hat{\theta}}$
with parameters $\theta_{GT}$ and $\hat{\theta}$,
respectively, and measure the discrepancy between the two transformed grids by summing the squared distances
between the corresponding grid points:
\begin{equation}
\mathcal{L}(\hat{\theta},\theta_{GT}) = \frac{1}{N} \sum_{i=1}^{N} \text{d}(\mathcal{T}_{\hat{\theta}}(g_i),\mathcal{T}_{\theta_{GT}}(g_i))^2
\end{equation}
\noindent where $\mathcal{G}=\{g_i\}=\{(x_i,y_i)\}$ is the uniform grid used, and $N=|\mathcal{G}|$. We define the grid as having $x_i,y_i \in \{s:s=-1+0.1\times n, n\in \{0,1,\dots,20\}\}$, that is to say, each coordinate belongs to a partition of $[-1,1]$ in equally spaced subintervals of steps $0.1$.
Note that we construct the coordinate system such that the center of the image is at $(0,0)$ and that the width and height of the image are equal to $2$, \ie\ the bottom left and top right corners have coordinates $(-1,-1)$ and $(1,1)$, respectively.

The gradient of the loss function with respect to the transformation parameters, needed to perform
backpropagation in order to learn network weights, can be computed easily if the location of the transformed
grid points $\mathcal{T}_{\hat{\theta}}(g_i)$ is differentiable with respect to $\hat{\theta}$.
This is commonly the case,
for example, when $\mathcal{T}$ is an affine transformation, $\mathcal{T}_{\hat{\theta}}(g_i)$ is linear in parameters $\hat{\theta}$ and therefore the loss can be differentiated in a straightforward manner.

\subsection{Training from synthetic transformations}
\label{sec:synth}

\begin{figure}
\centering \includegraphics[width=0.95\columnwidth]{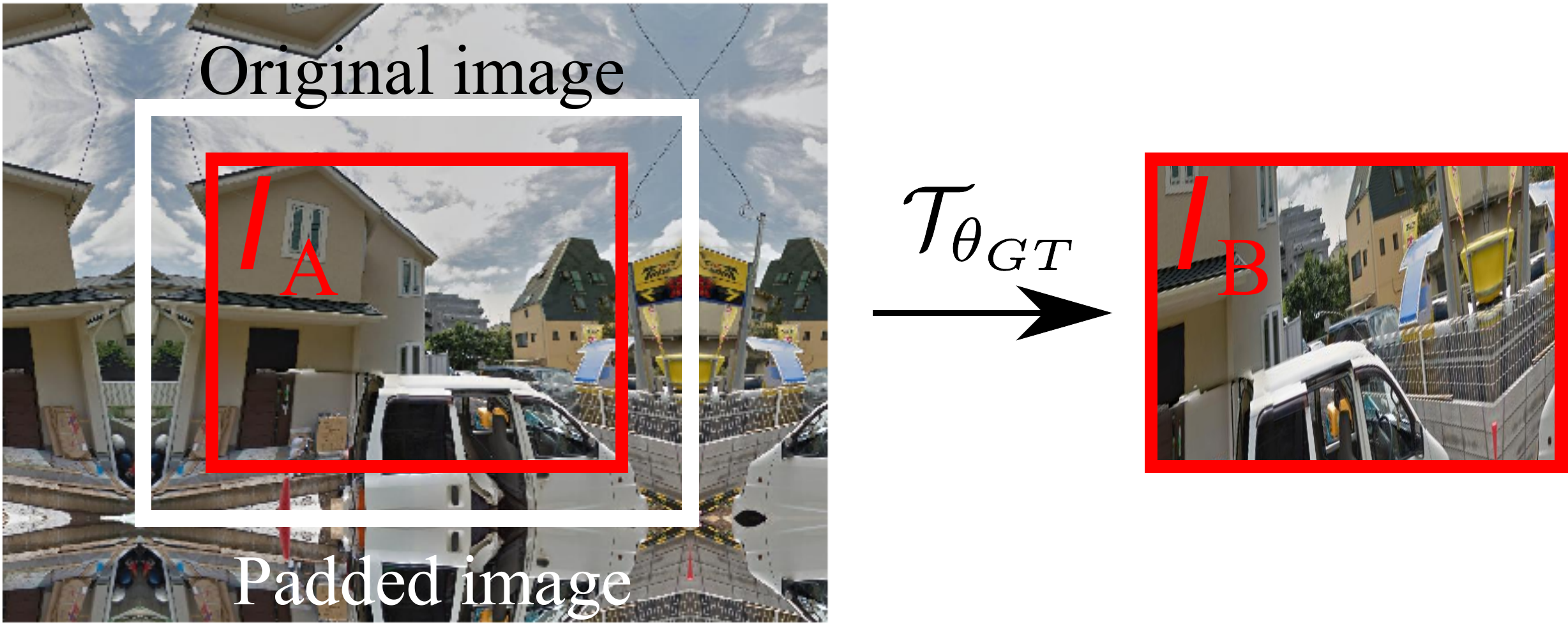} 
\captionsetup{font={small}}
\caption{{\bf Synthetic image generation.}
Symmetric padding is added to the original image to enlarge the sampling region, its central crop is used as image A,
and image B is created by performing a randomly sampled transformation $\mathcal{T}_{\theta_{GT}}$.
 } \label{fig:syntim}
 \vspace{-0.4cm}
\end{figure}

Our training procedure requires fully supervised training data consisting of image pairs and a known
geometric relation. Training CNNs usually requires a lot of data, and no public datasets exist that contain many image pairs annotated with their geometric transformation.
Therefore, we opt for training from synthetically generated data, which gives us the flexibility to
gather as many training examples as needed, for any 2-D geometric transformation of interest.
We generate each training pair $(I_A,I_B)$, by sampling 
$I_A$ from a public image dataset, and generating 
$I_B$
by applying a random transformation $\mathcal{T}_{\theta_{GT}}$ to 
$I_A$.
More precisely, 
$I_A$ is created from the central crop of the original image, while 
$I_B$ is created
by transforming the original image with added symmetrical padding in order to avoid border artifacts;
the procedure is shown in Fig.\,\ref{fig:syntim}.

\section{Experimental results}

In this section we describe our datasets, give implementation details, and compare our method to baselines and the state-of-the-art.
We also provide further insights into the components of our architecture.

\subsection{Evaluation dataset and performance measure}

Quantitative evaluation of our method is performed on the Proposal Flow dataset of Ham \etal \cite{ham2016}.
The dataset contains 900 image pairs depicting different instances of the same class, such as ducks and cars,
but with large intra-class variations, \eg\ the cars are often of different make, or the ducks can be of different subspecies. Furthermore, the images contain significant background clutter, as can be seen in Fig.\,\ref{fig:qualitative}.
The task is to predict the locations of predefined keypoints from image A in image B.
We do so by estimating a geometric transformation that warps image A into image B, and applying the same
transformation to the keypoint locations.
We follow the standard evaluation metric used for this benchmark, \ie\ the average probability of correct keypoint (PCK) \cite{Yang13}, being the proportion of keypoints that are correctly matched. A keypoint is considered to be matched correctly if its predicted location is within a distance of $\alpha\cdot\text{max}(h,w)$ of the target keypoint position, where $\alpha=0.1$ and $h$ and $w$ are the height and width of the object bounding box, respectively.

\subsection{Training data}

Two different training datasets for the affine and thin-plate spline stages, dubbed \emph{StreetView-synth-aff} and \emph{StreetView-synth-tps} respectively, were generated by applying synthetic transformations to images from the Tokyo Time Machine dataset \cite{arandjelovic2015netvlad} which contains Google Street View images of Tokyo.

Each synthetically generated dataset contains 40k images, divided into 20k for training and 20k for validation.
The ground truth transformation parameters were sampled independently from reasonable ranges, \eg\ for the affine transformation we sample the relative scale change of up to $2 \times$, while for thin-plate spline
we randomly jitter a $3 \times 3$ grid of control points by independently translating each point by up to one quarter of the image size in all directions.

In addition, a second training dataset for the affine stage was generated, created from the training set of Pascal VOC 2011 \cite{pascal-voc-2011} which we dubbed \emph{Pascal-synth-aff}. In Sec.\,\ref{sec:generalization}, we compare the performance of networks trained with  StreetView-synth-aff and Pascal-synth-aff and demonstrate the generalization capabilities of our approach.

\subsection{Implementation details}
We use the MatConvNet library \cite{vedaldi15matconvnet} and train the networks with stochastic gradient descent,
with learning rate $10^{-3}$, momentum 0.9, no weight decay and batch size of 16.
There is no need for jittering as instead of data augmentation we can simply generate more synthetic training data.
Input images are resized to $227 \times 227$ producing $15 \times 15$ feature maps that are passed into the matching layer. The affine and thin-plate spline stages are trained independently with the StreetView-synth-aff and StreetView-synth-tps datasets, respectively. 
Both stages are trained until convergence which typically occurs after 10 epochs, and takes 12 hours on a single GPU. 	
Our final method for estimating affine transformations uses an ensemble of two networks that independently regress the parameters, which are then averaged to produce the final affine estimate.
The two networks were trained on different ranges of affine transformations.
As in Fig.\,\ref{fig:hierarchy}, the estimated affine transformation is used to warp image A and pass it together
with image B to a 
second network which estimates the thin-plate spline transformation. All training and evaluation code, as well
as our trained networks, are online at~\cite{website}.

\subsection{Comparison to state-of-the-art}

We compare our method against SIFT Flow \cite{liu2011sift}, Graph-matching kernels (GMK) \cite{Duchenne11}, Deformable spatial pyramid matching (DSP) \cite{Kim13}, DeepFlow \cite{revaud2015deepmatching},
and all three variants of Proposal Flow (NAM, PHM, LOM) \cite{ham2016}.
As shown in Tab.\,\ref{tab:pck}, our method outperforms all others and sets
the new state-of-the-art on this data.
The best competing methods are based on Proposal Flow and make use of
object proposals, which enables them to guide the matching towards regions of images
that contain objects. Their performance varies significantly with the choice of the object proposal method, illustrating the
importance of this guided matching. On the contrary, our method does not use any guiding,
but it still manages to outperform even the best Proposal Flow and object proposal combination.

Furthermore, we also compare to affine transformations estimated with RANSAC  using the same descriptors as our method
(VGG-16 \texttt{pool4}). The parameters of this baseline have been tuned extensively to obtain the best result by adjusting the thresholds for
the second nearest neighbor test and by pruning proposal transformations which are
outside of the range of likely transformations.
Our affine estimator outperforms the RANSAC baseline on this task with 49\% (ours) compared to 47\% (RANSAC).

\begin{table}
\centering
\small
\begin{tabular}{lc}
\hline
Methods & PCK (\%) \\
\hline\hline
DeepFlow \cite{revaud2015deepmatching} & 20 \\
GMK \cite{Duchenne11} & 27 \\
SIFT Flow \cite{liu2011sift} & 38 \\
DSP \cite{Kim13} & 29 \\
Proposal Flow NAM \cite{ham2016} & 53 \\
Proposal Flow PHM \cite{ham2016} & 55 \\
Proposal Flow LOM \cite{ham2016} & 56 \\
RANSAC with our features (affine) & 47 \\
Ours (affine) & 49 \\
Ours (affine + thin-plate spline) & 56 \\
Ours (affine ensemble + thin-plate spline) & {\bf 57} \\
\hline \\
\end{tabular}
    \vspace{-0.4cm}
    \captionsetup{font={small}}
    \caption{{\bf Comparison to state-of-the-art and baselines.}
Matching quality on the Proposal Flow dataset measured in terms of PCK.
The Proposal Flow methods have four different PCK values, one for each of the four employed region proposal methods.
All the numbers apart from ours and RANSAC are taken from \cite{ham2016}.
}
    \label{tab:pck}
    \vspace{-0.3cm}
\end{table}

\subsection{Discussions and ablation studies}
In this section we examine the importance of various components of our architecture.
Apart from training on the StreetView-synth-aff dataset, we also train on Pascal-synth-aff
which contains images that are more similar in nature to the images in the
Proposal Flow benchmark.
The results of these ablation studies are summarized in Tab.\,\ref{tab:abltion}.

\begin{table}
\centering
\small 
\setlength\tabcolsep{3pt} 
\begin{tabular}{l@{}cc}
\hline
Methods & StreetView-synth-aff & Pascal-synth-aff \\
\hline\hline
Concatenation \cite{detone2016deep} & 26 & 29 \\
Subtraction \cite{kanazawa2016warpnet} & 18 & 21 \\
Ours without normalization & 44 & -- \\
Ours & {\bf 49} & {\bf 45} \\
\hline \\
\end{tabular}
\captionsetup{font={small}}
    \vspace{-0.4cm}
    \caption{{\bf Ablation studies.}
Matching quality on the Proposal Flow dataset measured in terms of PCK.
All methods use the same features (VGG-16 cropped at pool4).
The networks were trained on the StreetView-synth-aff and Pascal-synth-aff datasets.
For these experiments, only the affine transformation is estimated.
}
    \label{tab:abltion}
    \vspace{-0.3cm}
\end{table}

\begin{figure*}
    \centering
 \includegraphics[width=\linewidth]{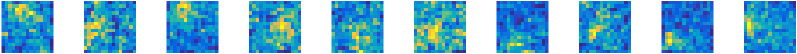}\vspace{-0.1cm}
 \captionsetup{font={small}} 
    \caption{{\bf Filter visualization.}
Some convolutional filters from the first layer of the regressor, acting on the tentative correspondence map,
show preferences to spatially co-located features that transform consistently to the other image,
thus learning to perform the local neighborhood consensus criterion often used in classical feature matching.
Refer to the text for more details on the visualization.}
    \label{fig:filters}
\end{figure*}

\begin{figure*}
    \centering
\setlength\tabcolsep{2.5pt} 
\begin{tabular}{cccc}
 \includegraphics[width=0.24\linewidth]{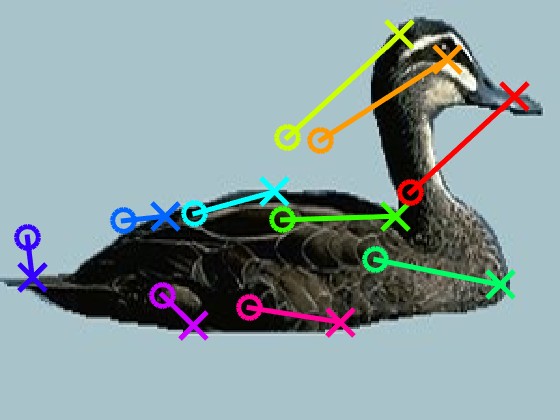}&
 \includegraphics[width=0.24\linewidth]{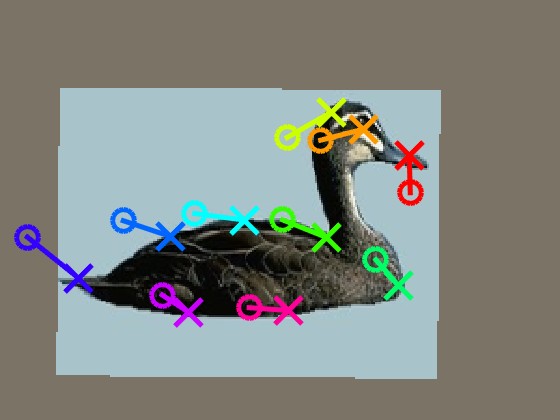}&
 \includegraphics[width=0.24\linewidth]{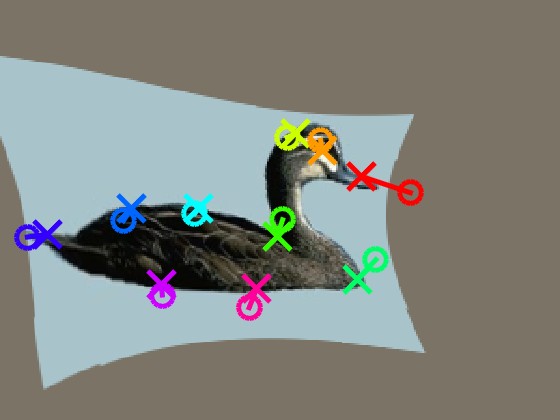}&
 \includegraphics[width=0.24\linewidth]{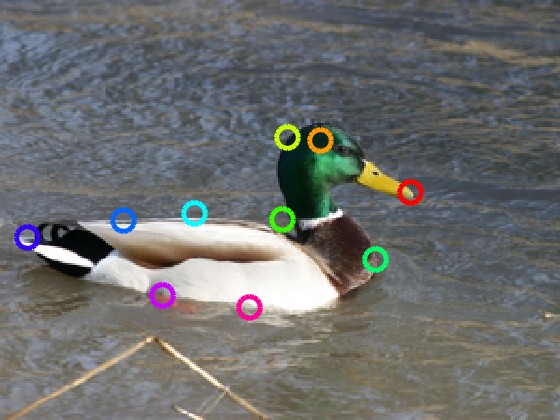}\\
 
 \includegraphics[width=0.24\linewidth]{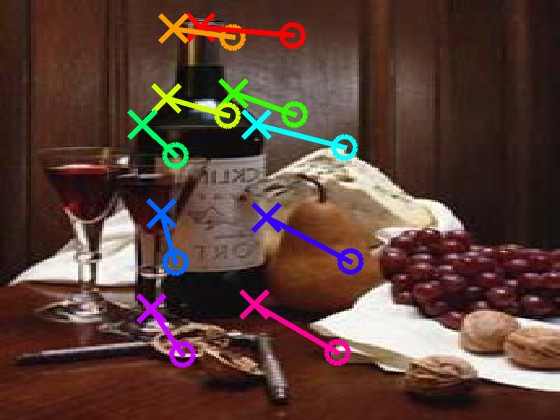}&
 \includegraphics[width=0.24\linewidth]{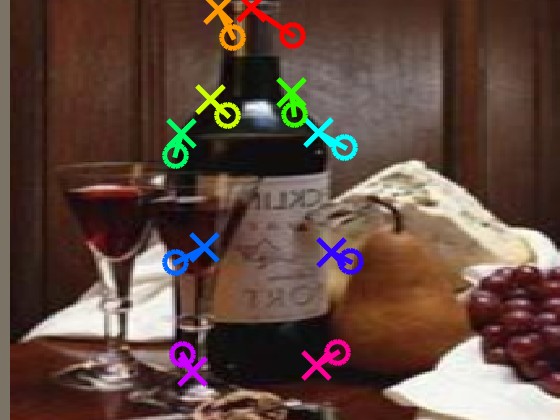}&
 \includegraphics[width=0.24\linewidth]{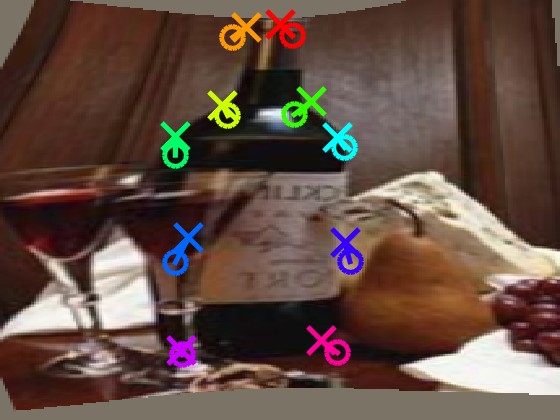}&
 \includegraphics[width=0.24\linewidth]{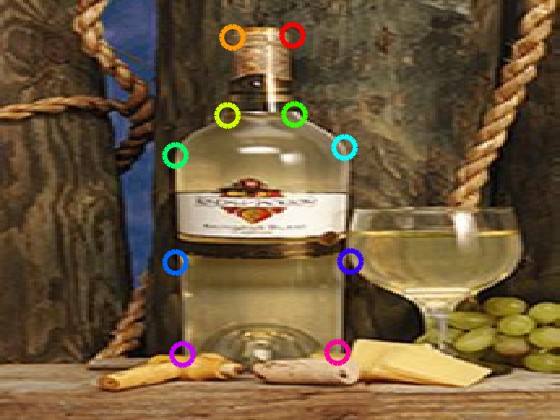}\\
 
 \includegraphics[width=0.24\linewidth]{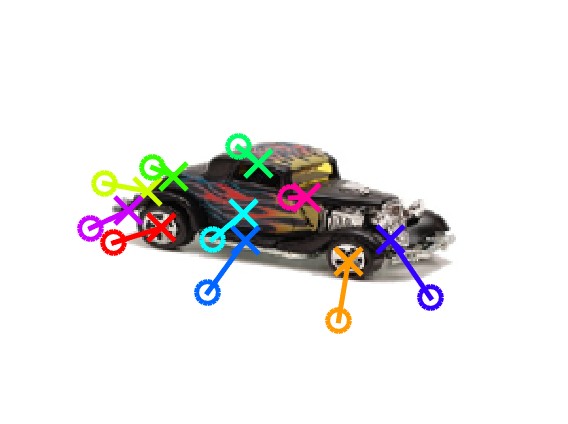}&
 \includegraphics[width=0.24\linewidth]{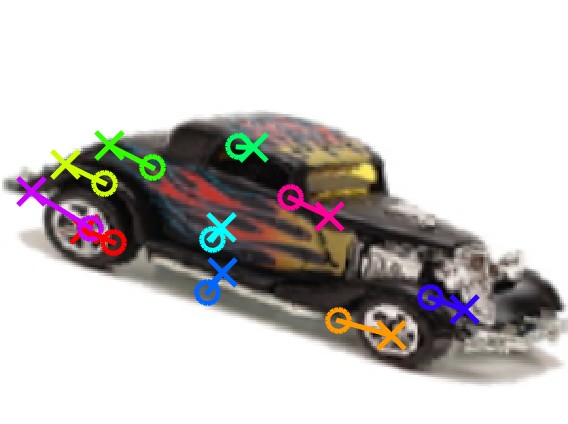}&
 \includegraphics[width=0.24\linewidth]{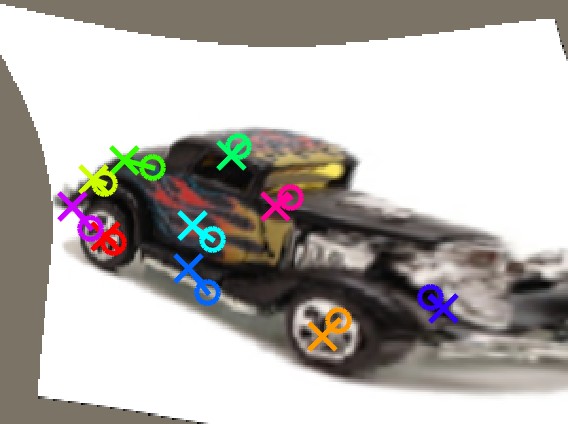}&
 \includegraphics[width=0.24\linewidth]{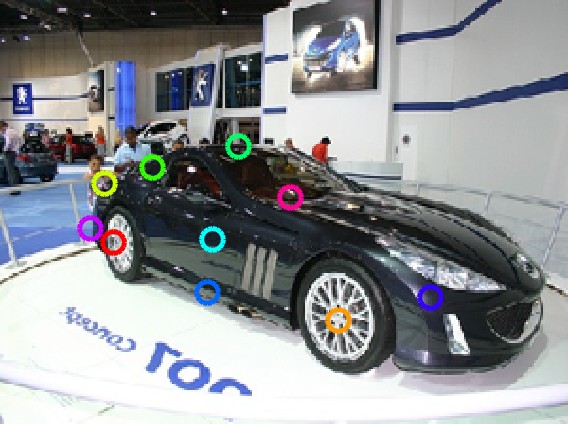}\\
 
 Image A&
 Aligned A (affine)&
 Aligned A (affine+TPS)&
 Image B
\end{tabular}
\captionsetup{font={small}}
    \caption{{\bf Qualitative results on the Proposal Flow dataset.}
Each row shows one test example from the Proposal Flow dataset.
Ground truth matching keypoints, only used for alignment evaluation, are depicted as crosses and circles for images A and B, respectively. Keypoints of same color are supposed to match each other after image A is aligned to image B. To illustrate the matching error, we also overlay keypoints of B onto different alignments of A so that lines that connect matching keypoints indicate the keypoint position error vector.
Our method manages to roughly align the images with an affine transformation (column 2), and then perform finer
alignment using thin-plate spline (TPS, column 3).
It successfully handles background clutter, translations, rotations, and large changes in appearance and scale, as well as
non-rigid transformations and some perspective changes.
Further examples are shown \isArXiv{in appendix \ref{apx:propflow_results}}{in the supplementary material~\cite{appendix}}.
}
    \label{fig:qualitative}
\end{figure*}

\begin{figure*}
    \centering
 \setlength\tabcolsep{3pt} 
 \begin{tabular}{ccccc}

 \includegraphics[width=0.19\linewidth]{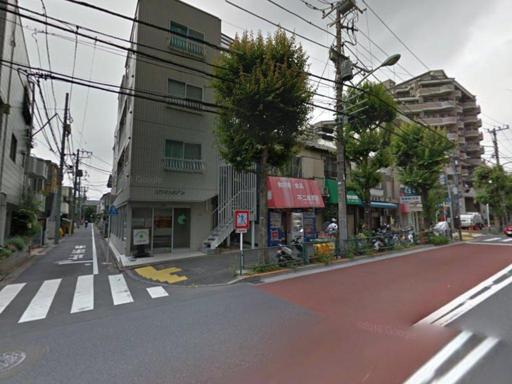}&
 \includegraphics[width=0.19\linewidth]{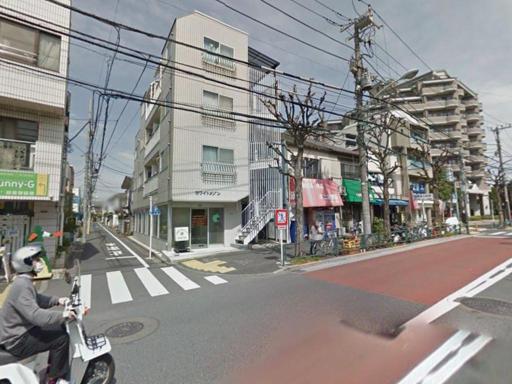}&
 \includegraphics[width=0.19\linewidth]{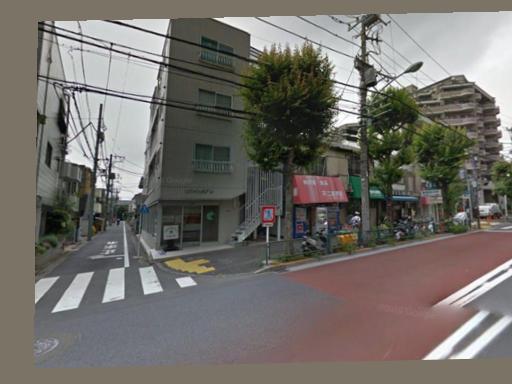}&
 \includegraphics[width=0.19\linewidth]{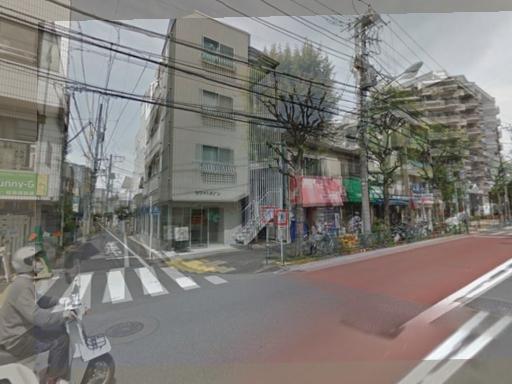}&
 \includegraphics[width=0.19\linewidth]{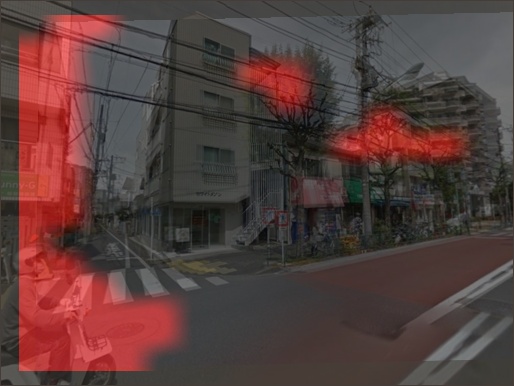}\\
 
 \includegraphics[width=0.19\linewidth]{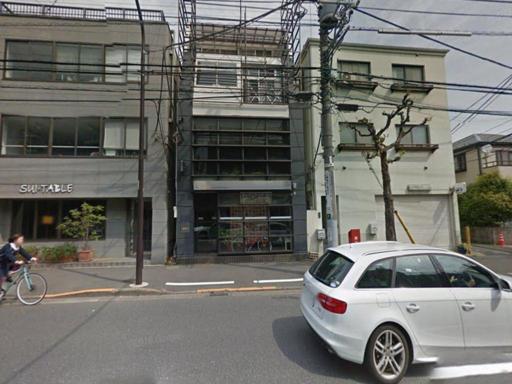}&
 \includegraphics[width=0.19\linewidth]{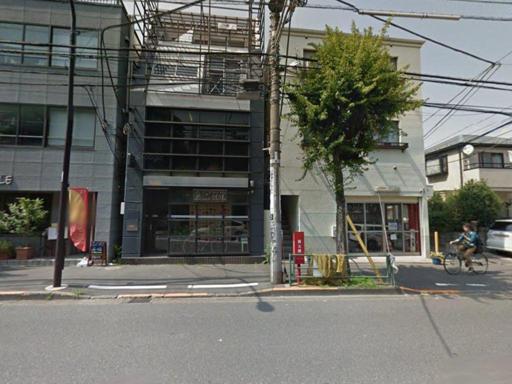}&
 \includegraphics[width=0.19\linewidth]{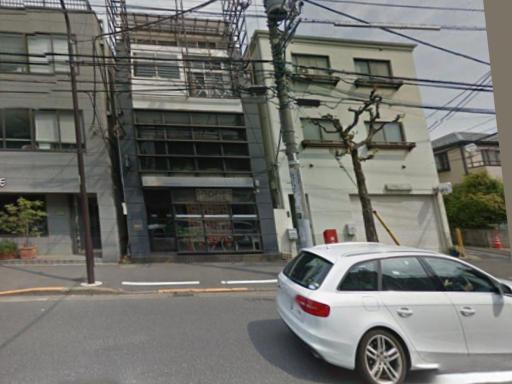}&
 \includegraphics[width=0.19\linewidth]{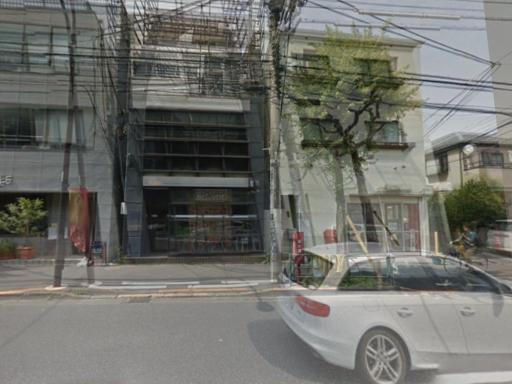}&
 \includegraphics[width=0.19\linewidth]{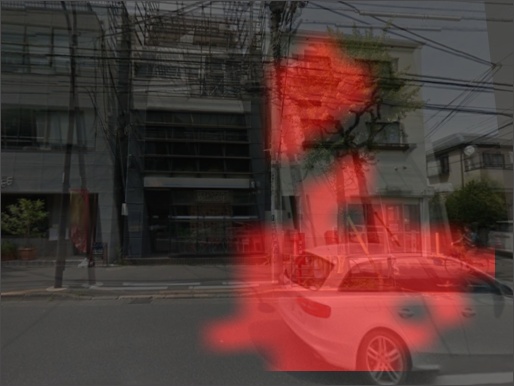}\\

 {\small (a) Image A}&
 {\small (b) Image B}&
 {\small (c) Aligned image A}&
 {\small (d) Overlay of (b) and (c)}&
 {\small (e) Difference map}
\end{tabular}
\captionsetup{font={small}}
    \caption{{\bf Qualitative results on the Tokyo Time Machine dataset.}
Each row shows a pair of images from the Tokyo Time Machine dataset, and our alignment along with a ``difference map'', highlighting absolute differences between aligned images in the descriptor space.
Our method successfully aligns image A to image B despite 
of  
viewpoint and
scene changes (highlighted in the difference map).
}
    \label{fig:qualstreetview}
    \vspace{-0.3cm}
\end{figure*}

\paragraph{Correlation versus concatenation and subtraction.}
Replacing our correlation-based matching layer with feature concatenation or subtraction,
as proposed in \cite{detone2016deep} and \cite{kanazawa2016warpnet}, respectively,
incurs a large performance drop.
The behavior is expected as we designed the matching layer
to only keep information on pairwise descriptor similarities rather than
the descriptors themselves, as is good practice in classical geometry estimation methods,
while concatenation and subtraction do not follow this principle.

\paragraph{Generalization.}
\label{sec:generalization}
As seen in Tab.\,\ref{tab:abltion}, our method is relatively unaffected by the choice of training data
as its performance is similar regardless whether it was trained with StreetView or Pascal images.
We also attribute this to the design choice of operating on pairwise descriptor similarities rather than
the raw descriptors.

\paragraph{Normalization.}
Tab.\,\ref{tab:abltion} also shows the importance of the correlation map normalization step,
where the normalization improves results from 44\% to 49\%.
The step mimics the second nearest neighbor test used in classical feature matching \cite{lowe2004distinctive}, as discussed in Sec.\,\ref{sec:fusion}.
Note that \cite{fischer2015flownet} also uses a correlation layer, but they do not normalize
the map in any way, which is clearly suboptimal.

\paragraph{What is being learned?}
\label{sec:whatislearned}
We examine filters from the first convolutional layer of the regressor,
which operate directly on the output of the matching layer, \ie\ the tentative correspondence map.
Recall that each spatial location in the correspondence map (see Fig.\,\ref{fig:correlation}, in green) contains all similarity scores between that
feature in image B and all features in image A.
Thus, each single 1-D slice through the weights of one convolutional filter at a particular spatial location
can be visualized as an image, showing filter's preferences to features in image B that match to specific
locations in image A. For example, if the central slice of a filter contains all zeros apart from a peak at the top-left
corner, this filter responds positively to features in image B that match to the top-left of image A.
Similarly, if many spatial locations of the filter produce similar visualizations, then this filter is highly sensitive
to spatially co-located features in image B that all match to the top-left of image A.
For visualization, we pick the peaks from all slices of filter weights and average them together
to produce a single image. Several filters shown in Fig.\,\ref{fig:filters} confirm our hypothesis
that this layer has learned to mimic local neighborhood consensus as some filters
respond strongly to spatially co-located features in image B that match to spatially consistent locations in image A.
Furthermore, it can be observed that the size of the preferred spatial neighborhood varies across filters, thus showing that the filters are discriminative of the scale change.

\subsection{Qualitative results}

Fig.\,\ref{fig:qualitative} illustrates
the effectiveness of our method in category-level matching, where challenging pairs of images
from the Proposal Flow dataset \cite{ham2016}, containing large intra-class variations, are aligned correctly.
The method is able to robustly, in the presence of clutter, estimate
large translations, rotations, scale changes, as well as non-rigid transformations
and some perspective changes. Further examples are shown \isArXiv{in appendix \ref{apx:propflow_results}}{in the supplementary material~\cite{appendix}}.

Fig.\,\ref{fig:qualstreetview} shows the quality of instance-level matching, where different
images of the same scene are aligned correctly.
The images are taken from the Tokyo Time Machine dataset \cite{arandjelovic2015netvlad}
and are captured at different points in time which are months or years apart.
Note that, by automatically highlighting the differences (in the feature space) between the aligned images,
it is possible to detect changes in the scene, such as occlusions, changes in vegetation,
or structural differences \eg\ new buildings being built.

\section{Conclusions}

We have described a network architecture for geometric matching fully trainable from synthetic imagery without the need for manual annotations. Thanks to our matching layer, the network generalizes well to never seen before imagery, reaching state-of-the-art results on the challenging Proposal Flow dataset for category-level matching. This opens-up the possibility of applying our architecture to other difficult correspondence problems such as matching across large changes in illumination (day/night)~\cite{arandjelovic2015netvlad} or depiction style~\cite{Aubry13}.
 
{\small
\paragraph{Acknowledgements.} 
This work has been partly supported by ERC grant LEAP (no.\ 336845), ANR project Semapolis (ANR-13-CORD-0003), the Inria CityLab IPL, CIFAR Learning in Machines $\&$ Brains program and ESIF, OP Research, development and education Project IMPACT No.\ CZ$.02.1.01/0.0/0.0/15\_003/0000468$.
}

{\small
\bibliographystyle{ieee}
\bibliography{shortstrings,egbib}
}

\isArXiv{\appendix

\section*{Appendices}

In these appendices we show additional qualitative results on the Proposal Flow dataset (appendix \ref{apx:propflow_results}), results on the Caltech-101 dataset~\cite{fei2006one} previously used for alignment in~\cite{ham2016} (appendix \ref{apx:caltech_results}), and details of our thin-plate spline (TPS) transformation model (appendix \ref{apx:tps}) used in the second stage to refine the affine transformation estimated in the first stage.

\section{Additional results on Proposal Flow dataset} \label{apx:propflow_results}
In figures \ref{fig:PF-large-scale}, \ref{fig:PF-large-viewpoint} and \ref{fig:PF-clutter} we show additional results of our method applied on image pairs from the Proposal Flow dataset \cite{ham2016}. 

Each row shows one test pair, where ground truth matching keypoints, only used for alignment evaluation, are depicted as crosses and circles for images A and B, respectively. Keypoints of the same color correspond to the same object parts and are supposed to match at the exact same position after image A is aligned to image B. To illustrate the alignment error we overlay keypoints of B onto different transformations of A so that lines that connect matching keypoints indicate the keypoint position error vector. 

Our method roughly aligns the two input images with an affine transformation (column 2), and then performs finer alignment using a thin-plate spline transformation (column 3).

These results demonstrate the ability of our network to successfully handle various challenging cases, such as changes in object scale (Fig.\ \ref{fig:PF-large-scale}), camera viewpoint changes (Fig.\ \ref{fig:PF-large-viewpoint}), and background clutter (Fig.\ \ref{fig:PF-clutter}). 

\paragraph{Limitations of the method.} In Fig. \ref{fig:PF-difficult} we show several difficult examples where our method is able to only partially align the two images. These cases tend to occur when there is a significant change of viewpoint, pose or scale, combined with a strong change in appearance (Fig. \ref{fig:PF-difficult}, examples 1-3). Furthermore, heavy clutter in both images can still present a challenge for the method (Fig. \ref{fig:PF-difficult}, example 4).

\begin{figure*}[!htbp]
    \vspace*{5mm}
    \centering
\setlength\tabcolsep{3pt} 
\begin{tabular}{cccc}
 \includegraphics[width=0.24\linewidth]{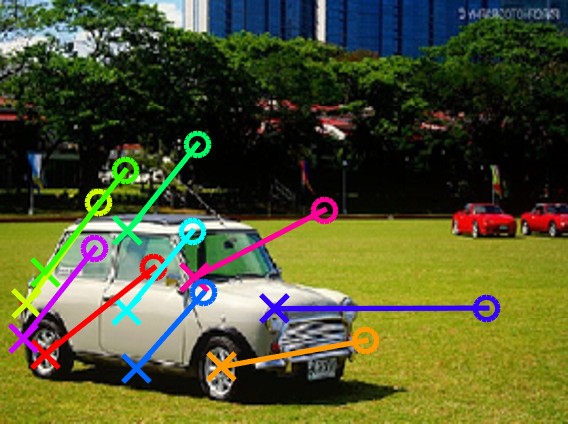}&
 \includegraphics[width=0.24\linewidth]{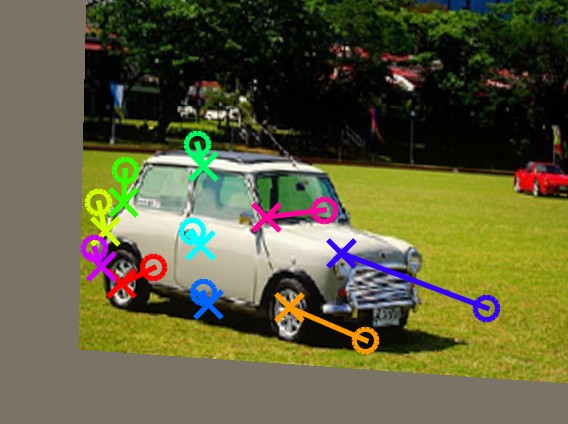}&
 \includegraphics[width=0.24\linewidth]{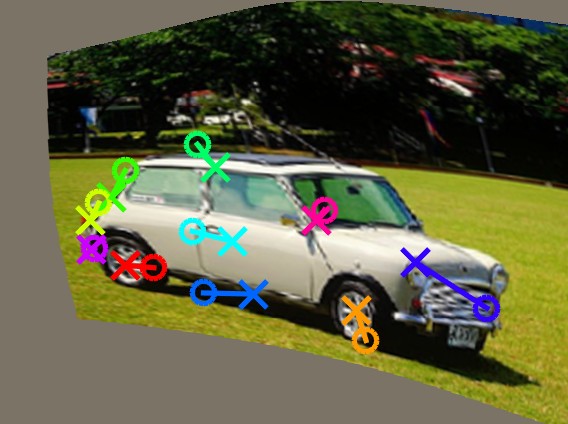}&
 \includegraphics[width=0.24\linewidth]{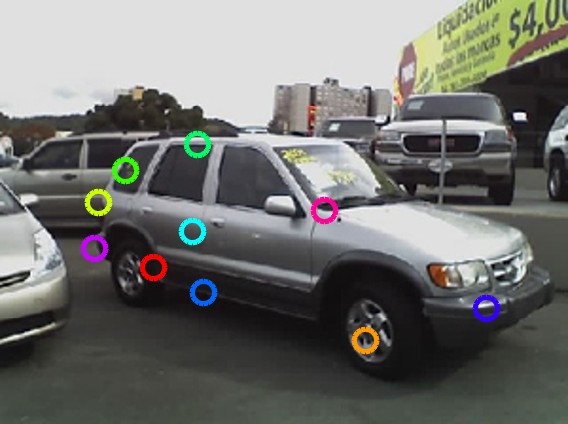}\\
 
 \includegraphics[width=0.24\linewidth]{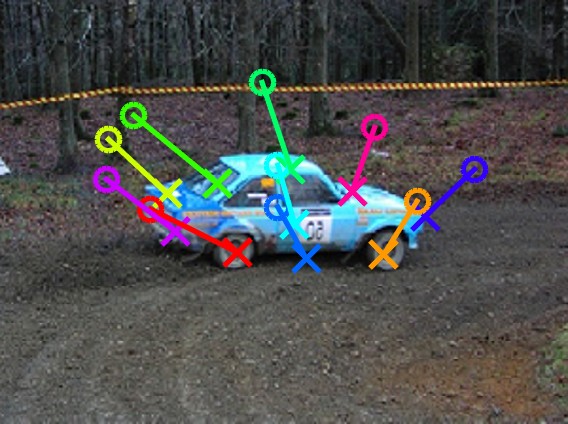}&
 \includegraphics[width=0.24\linewidth]{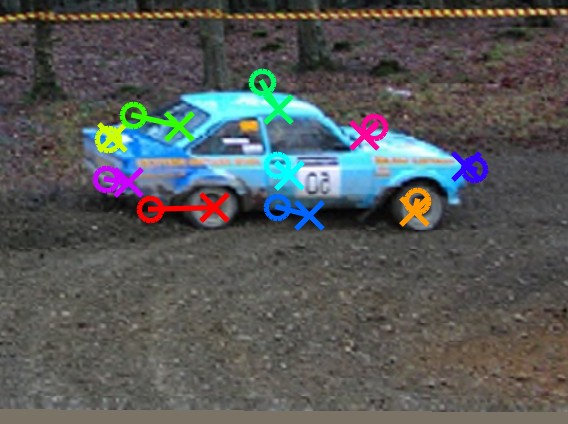}&
 \includegraphics[width=0.24\linewidth]{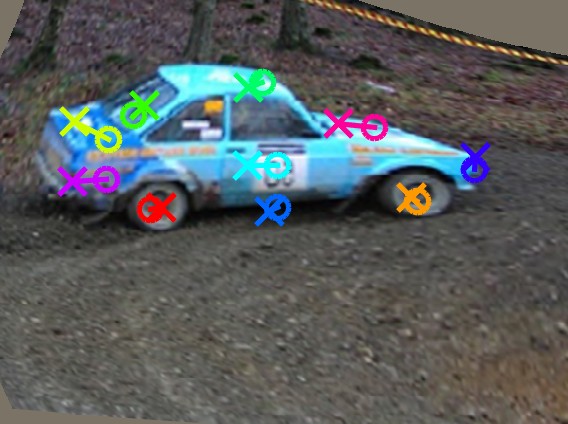}&
 \includegraphics[width=0.24\linewidth]{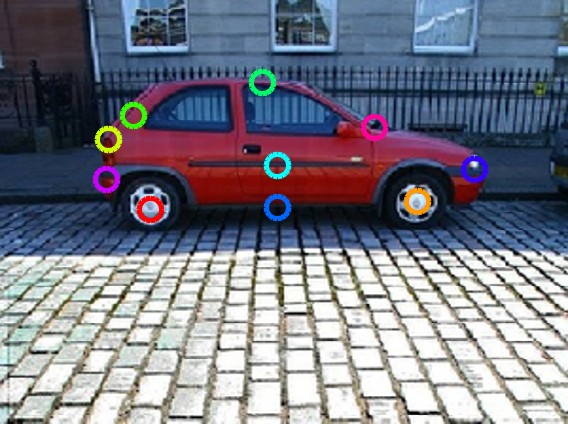}\\

 \includegraphics[width=0.24\linewidth]{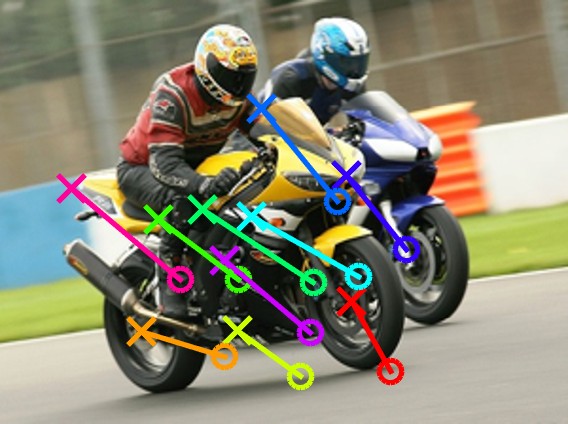}&
 \includegraphics[width=0.24\linewidth]{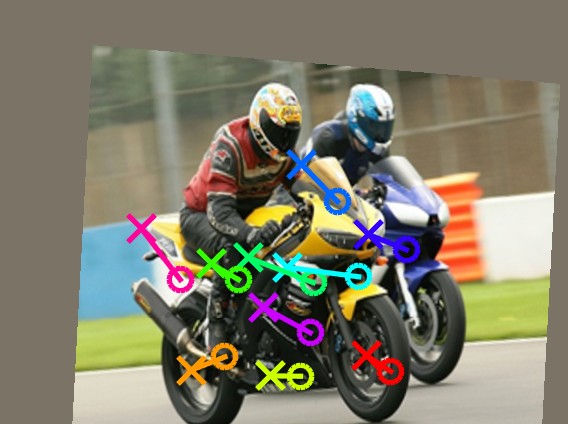}&
 \includegraphics[width=0.24\linewidth]{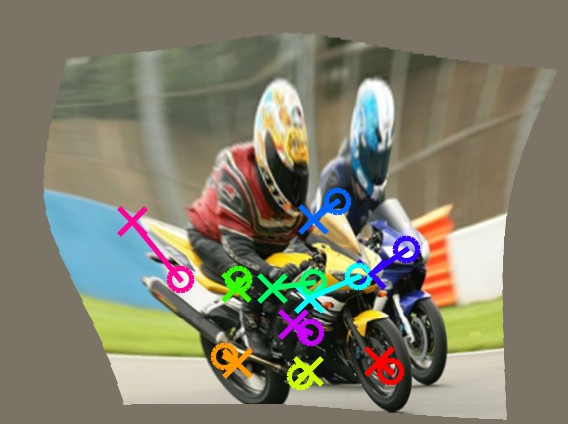}&
 \includegraphics[width=0.24\linewidth]{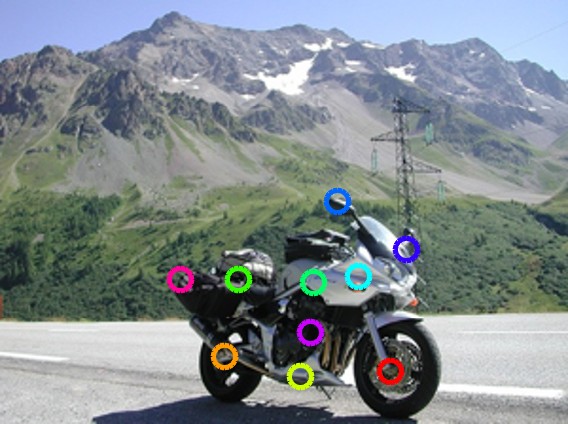}\\
 
 \includegraphics[width=0.24\linewidth]{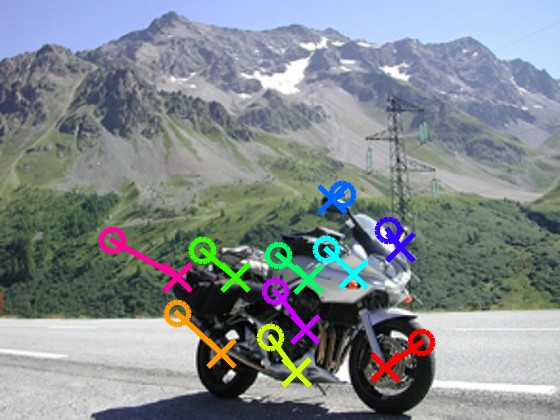}&
 \includegraphics[width=0.24\linewidth]{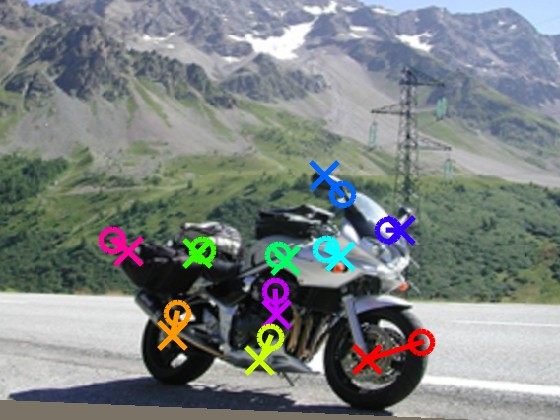}&
 \includegraphics[width=0.24\linewidth]{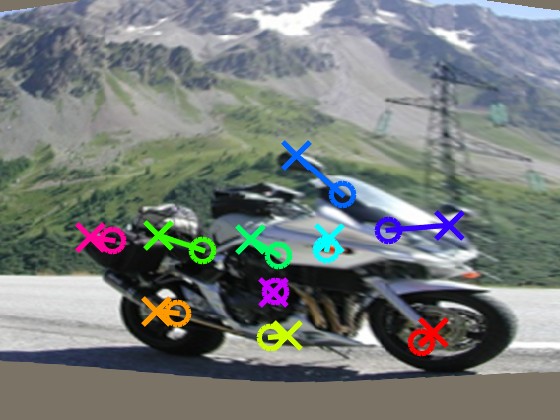}&
 \includegraphics[width=0.24\linewidth]{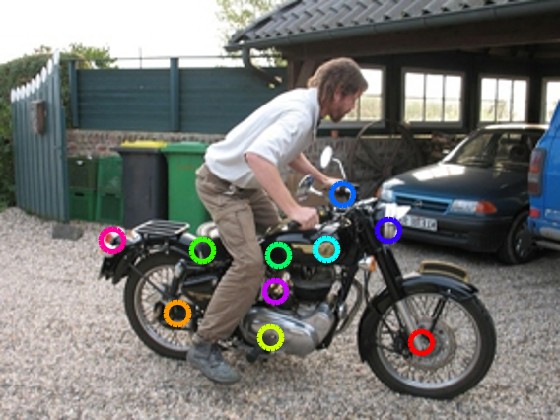}\\

 \includegraphics[width=0.24\linewidth]{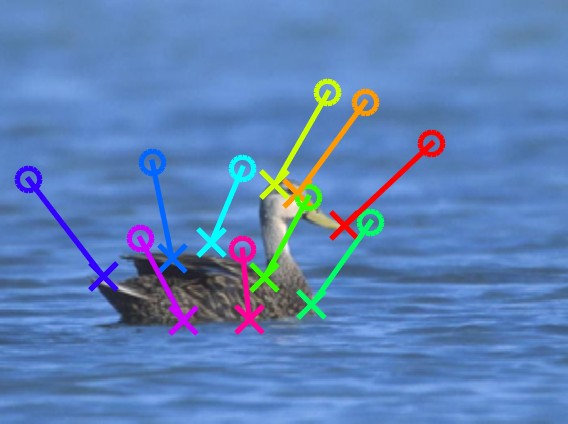}&
 \includegraphics[width=0.24\linewidth]{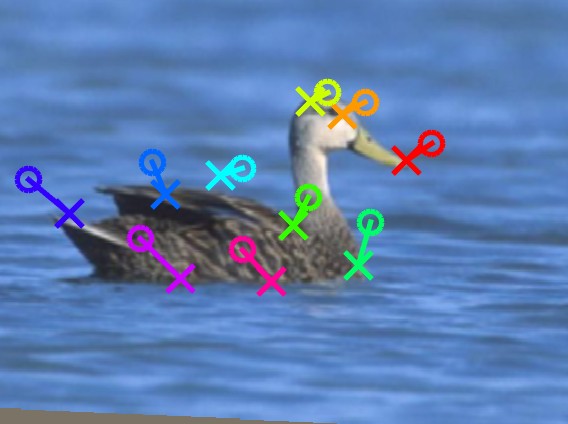}&
 \includegraphics[width=0.24\linewidth]{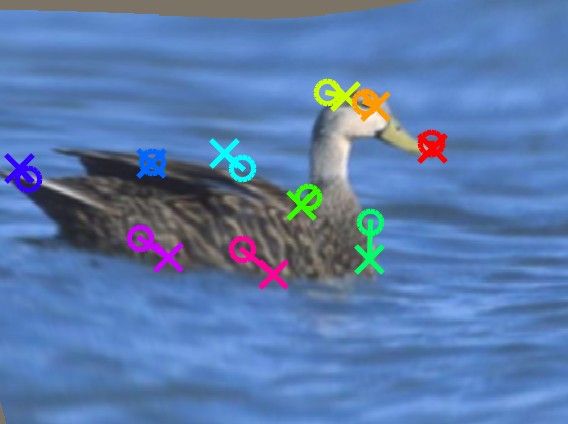}&
 \includegraphics[width=0.24\linewidth]{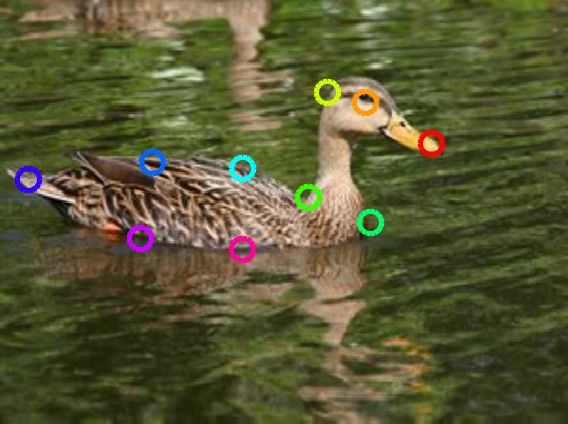}\\
 
 \includegraphics[width=0.24\linewidth]{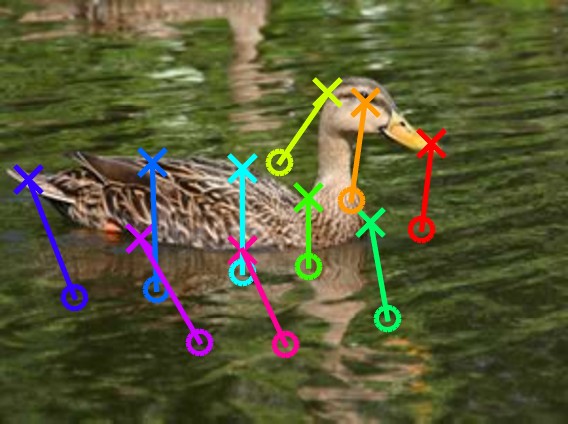}&
 \includegraphics[width=0.24\linewidth]{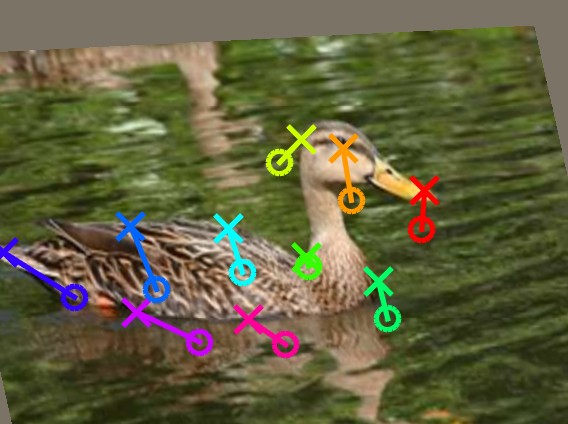}&
 \includegraphics[width=0.24\linewidth]{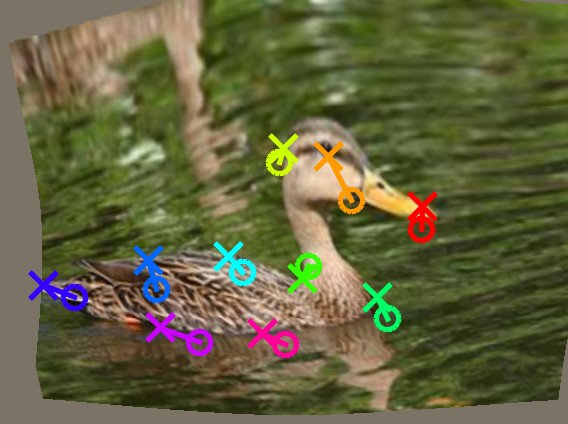}&
 \includegraphics[width=0.24\linewidth]{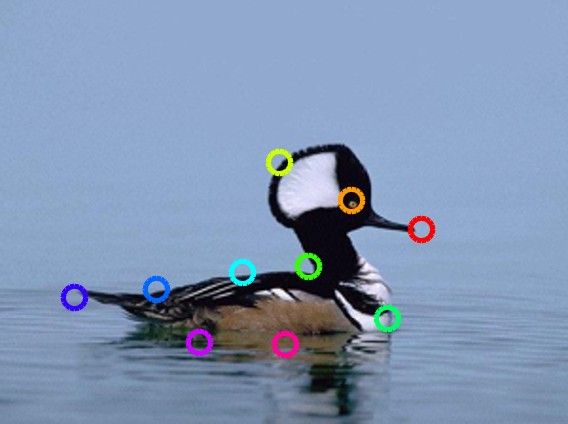}\\
 
 Image A&
 Aligned A (affine)&
 Aligned A (affine+TPS)&
 Image B
\end{tabular}
    \caption{{\bf Example image pairs from the Proposal Flow dataset featuring a significant change in scale.}}\label{fig:PF-large-scale}
    \vspace{4mm}
\end{figure*}

\begin{figure*}
    \centering
\setlength\tabcolsep{3pt} 
\begin{tabular}{cccc}
 \includegraphics[width=0.24\linewidth]{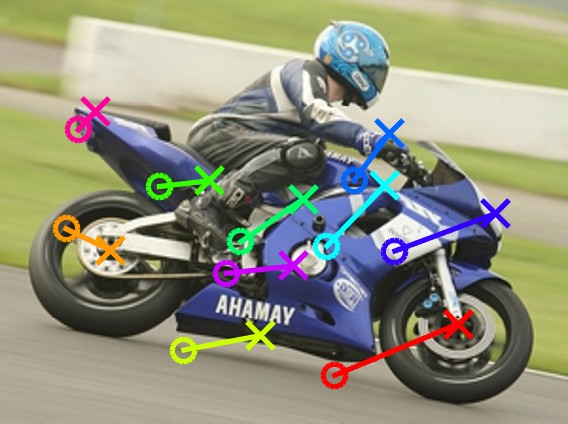}&
 \includegraphics[width=0.24\linewidth]{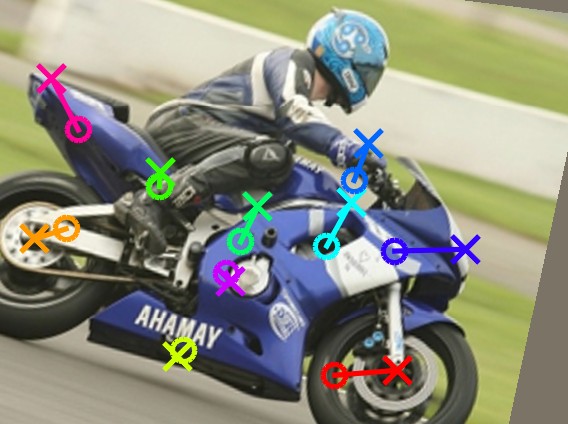}&
 \includegraphics[width=0.24\linewidth]{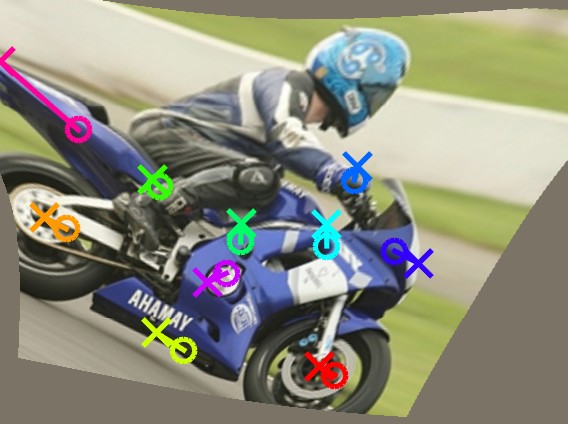}&
 \includegraphics[width=0.24\linewidth]{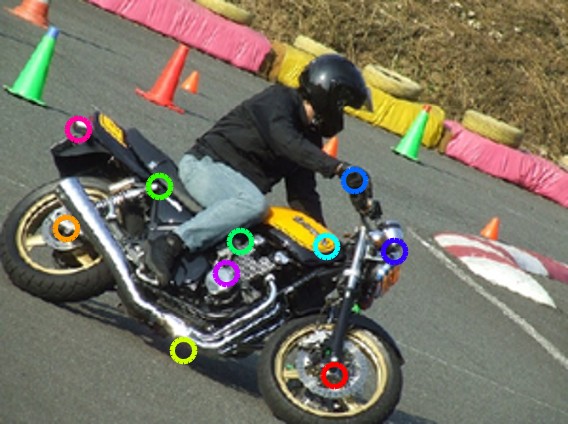}\\
 
 \includegraphics[width=0.24\linewidth]{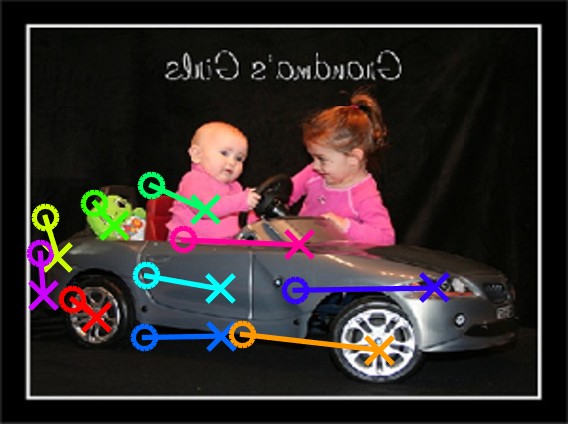}&
 \includegraphics[width=0.24\linewidth]{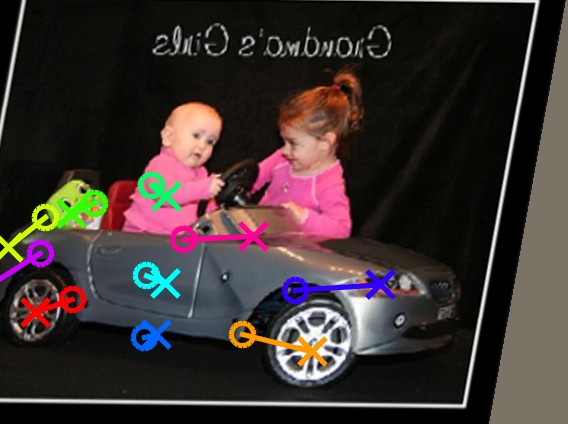}&
 \includegraphics[width=0.24\linewidth]{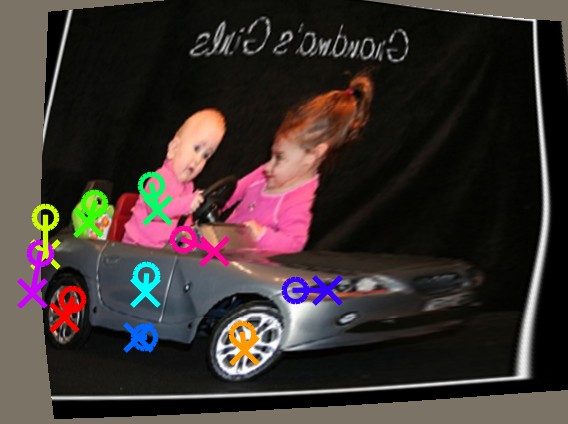}&
 \includegraphics[width=0.24\linewidth]{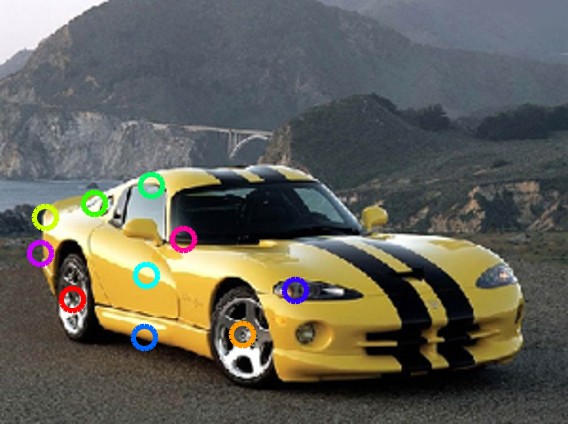}\\
 
 \includegraphics[width=0.24\linewidth]{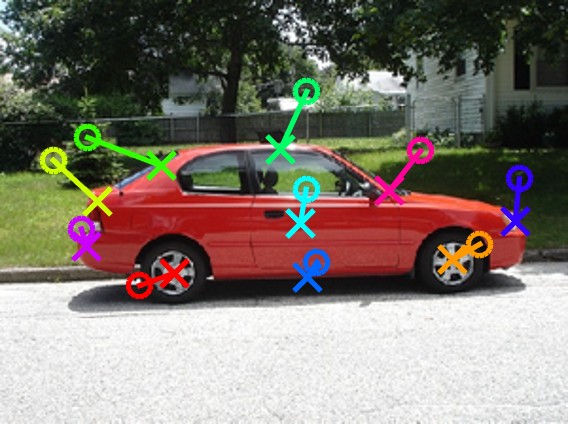}&
 \includegraphics[width=0.24\linewidth]{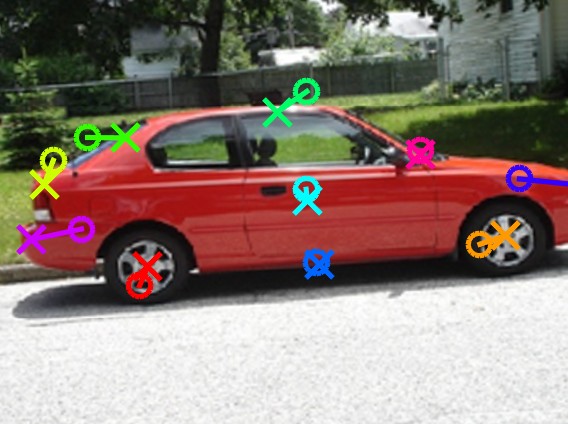}&
 \includegraphics[width=0.24\linewidth]{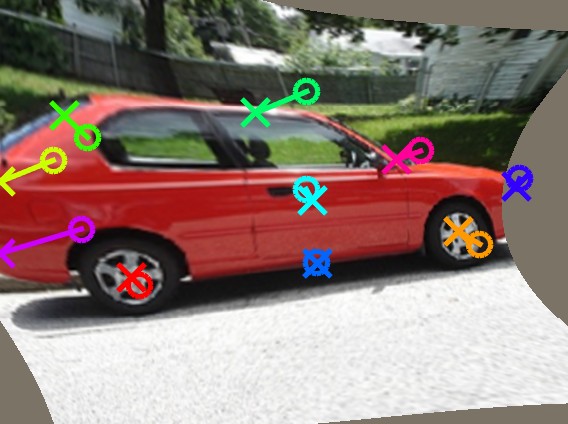}&
 \includegraphics[width=0.24\linewidth]{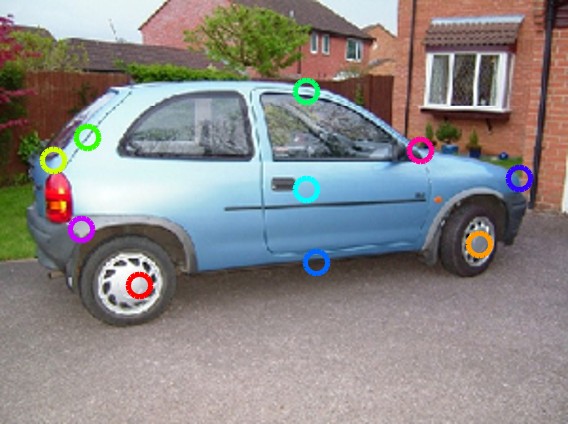}\\
 
 Image A&
 Aligned A (affine)&
 Aligned A (affine+TPS)&
 Image B
\end{tabular}\vspace{-2mm}
    \caption{{\bf Example image pairs from the Proposal Flow dataset demonstrating changes in viewpoint.} While the affine stage can correct for the size and orientation of the object, the thin-plate spline stage is able to perform a non-rigid deformation which can, to some extent, compensate for the change in viewpoint.}\label{fig:PF-large-viewpoint}
\vspace{4mm}
    \centering
\setlength\tabcolsep{3pt} 
\begin{tabular}{cccc}
 \includegraphics[width=0.24\linewidth]{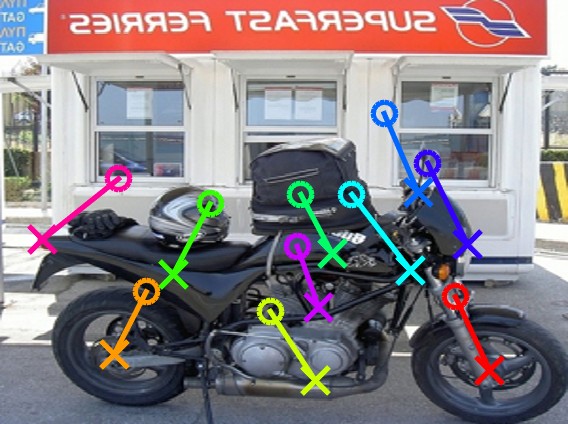}&
 \includegraphics[width=0.24\linewidth]{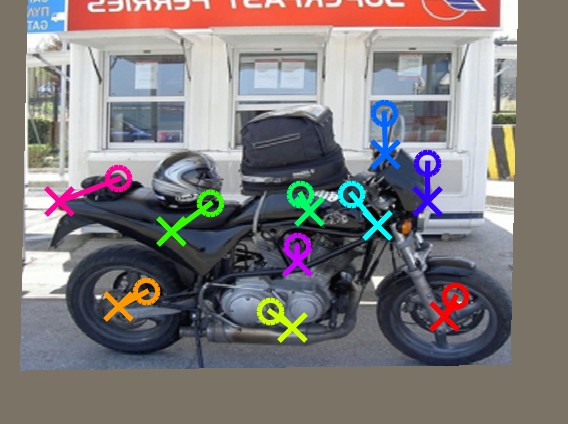}&
 \includegraphics[width=0.24\linewidth]{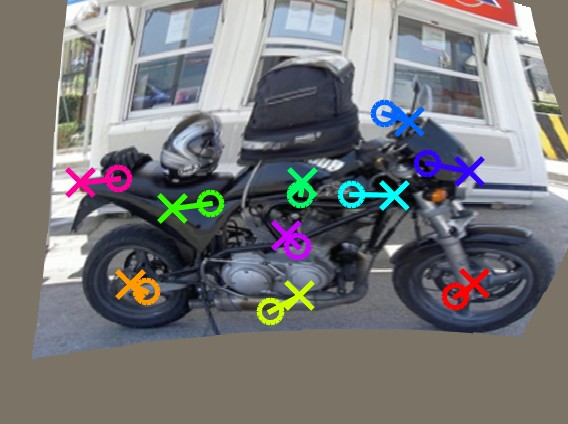}&
 \includegraphics[width=0.24\linewidth]{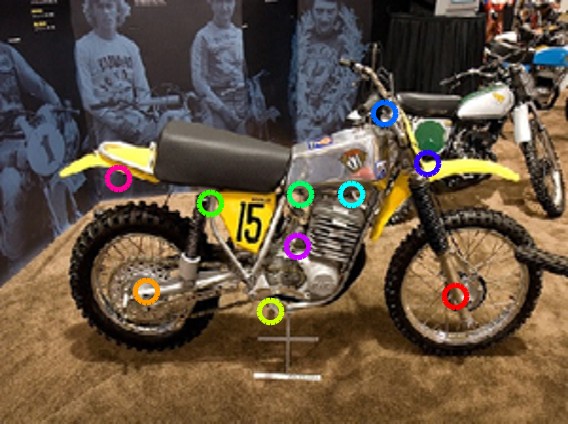}\\
 
 \includegraphics[width=0.24\linewidth]{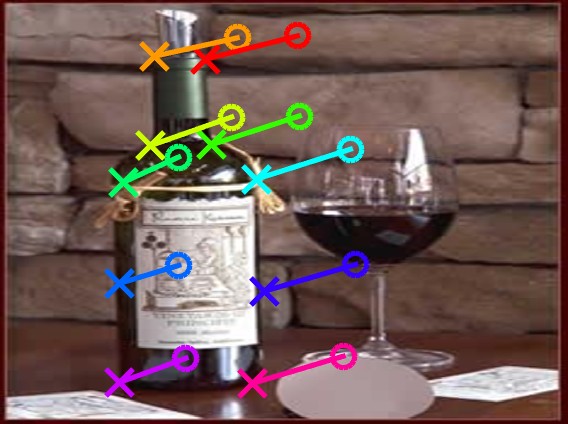}&
 \includegraphics[width=0.24\linewidth]{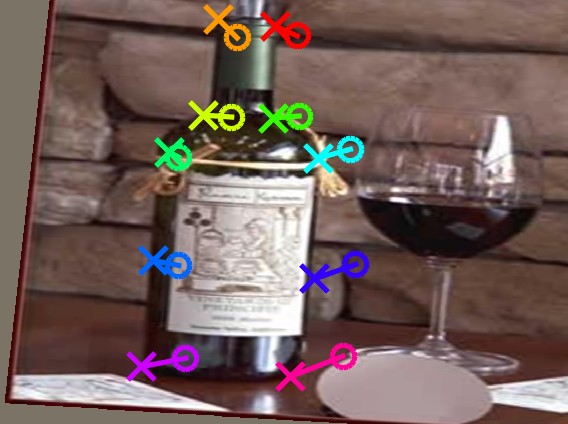}&
 \includegraphics[width=0.24\linewidth]{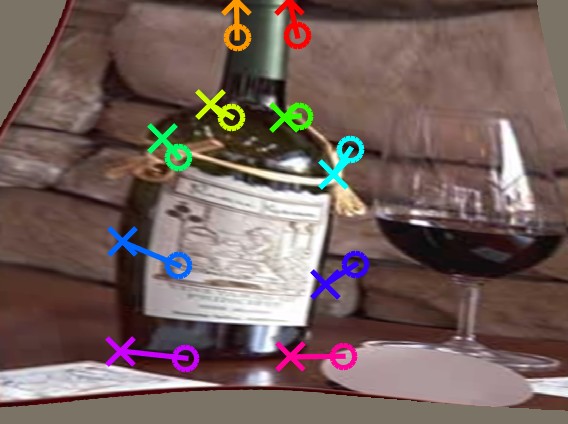}&
 \includegraphics[width=0.24\linewidth]{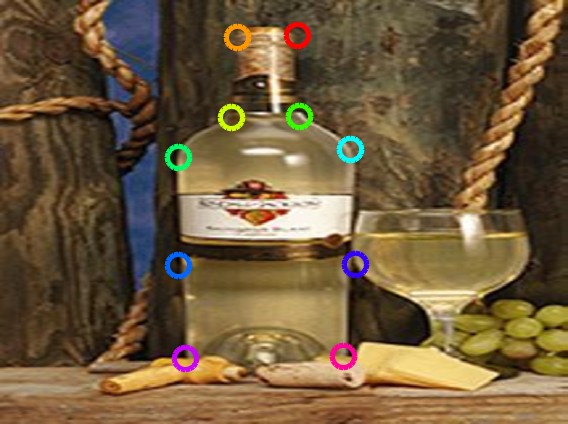}\\
 
 \includegraphics[width=0.24\linewidth]{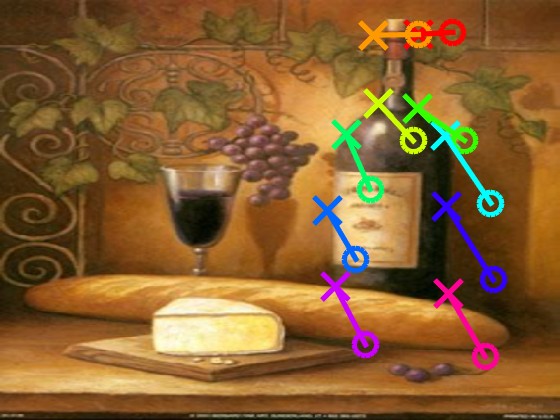}&
 \includegraphics[width=0.24\linewidth]{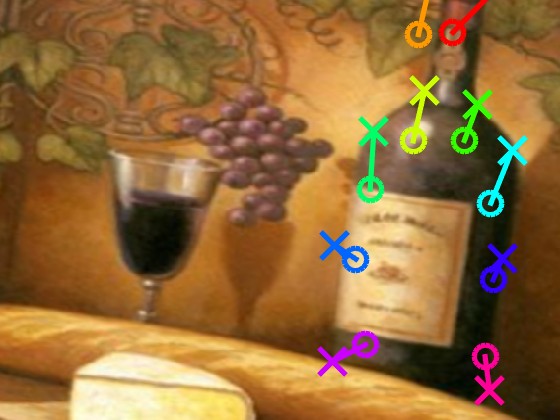}&
 \includegraphics[width=0.24\linewidth]{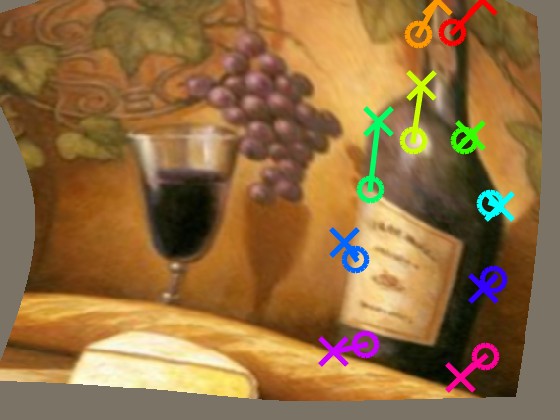}&
 \includegraphics[width=0.24\linewidth]{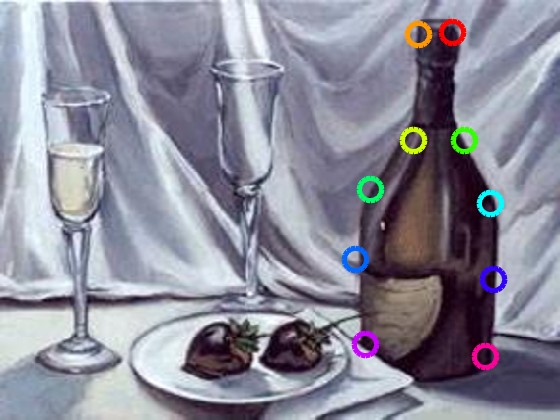}\\
 
 Image A&
 Aligned A (affine)&
 Aligned A (affine+TPS)&
 Image B
\end{tabular}\vspace{-2mm}
    \caption{{\bf Example image pairs from the Proposal Flow dataset with significant amount of background clutter.}}\label{fig:PF-clutter}
\end{figure*}

\begin{figure*}
    \centering
\setlength\tabcolsep{3pt} 
\begin{tabular}{cccc}
 \includegraphics[width=0.24\linewidth]{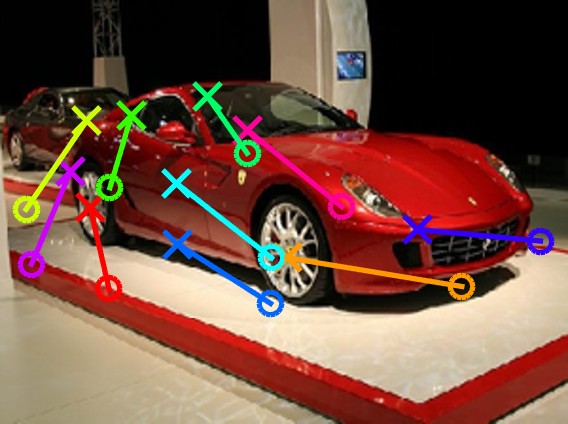}&
 \includegraphics[width=0.24\linewidth]{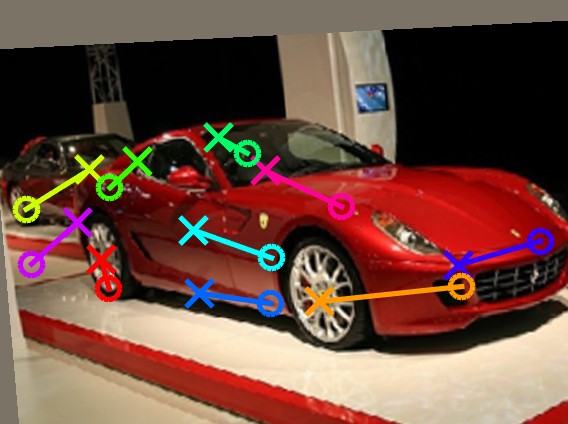}&
 \includegraphics[width=0.24\linewidth]{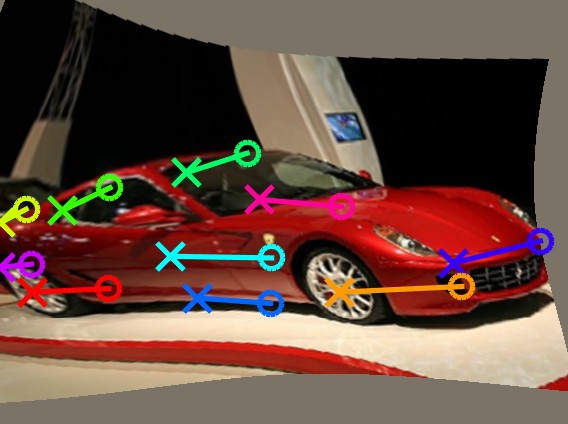}&
 \includegraphics[width=0.24\linewidth]{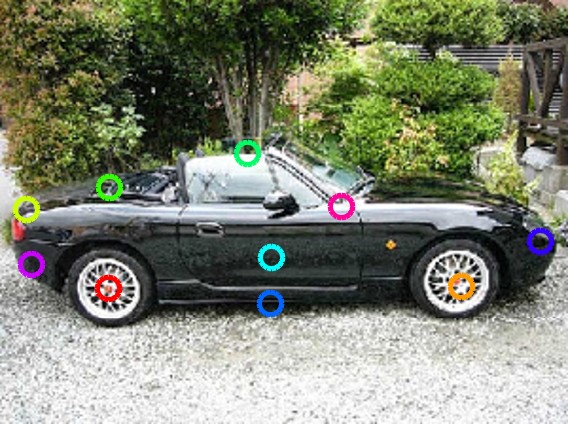}\\
 
 \includegraphics[width=0.24\linewidth]{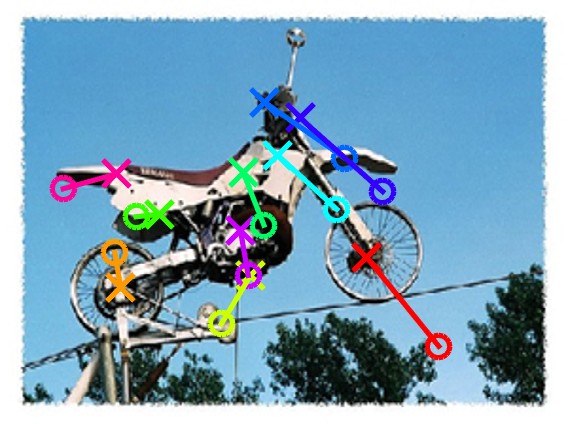}&
 \includegraphics[width=0.24\linewidth]{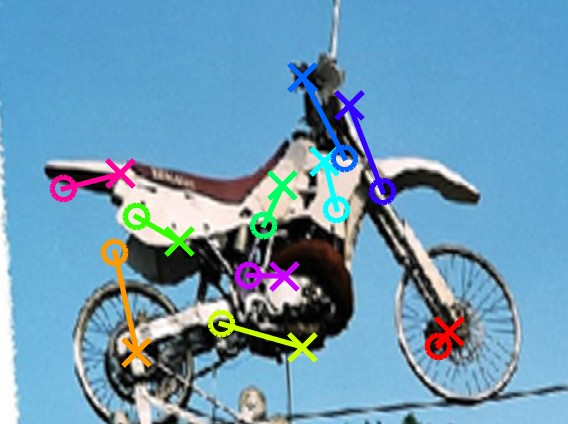}&
 \includegraphics[width=0.24\linewidth]{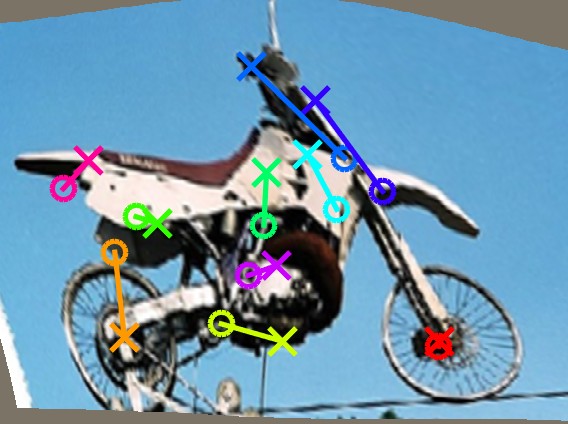}&
 \includegraphics[width=0.24\linewidth]{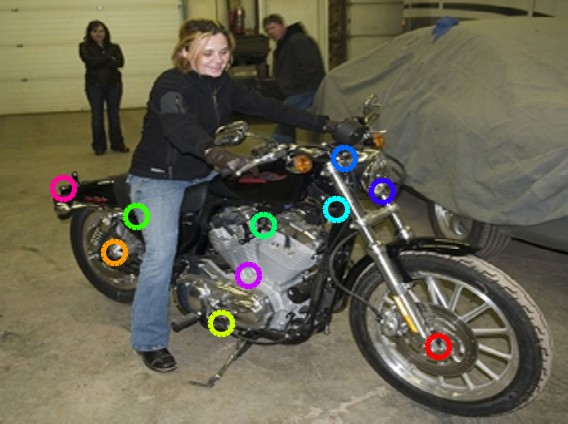}\\
 
 \includegraphics[width=0.24\linewidth]{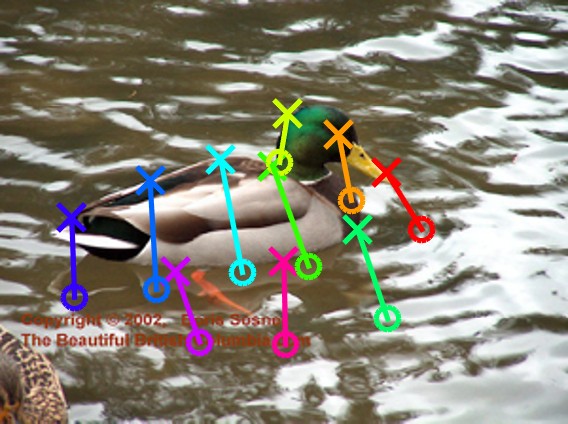}&
 \includegraphics[width=0.24\linewidth]{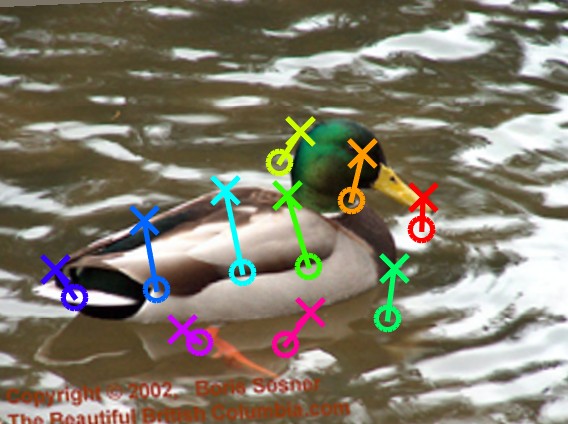}&
 \includegraphics[width=0.24\linewidth]{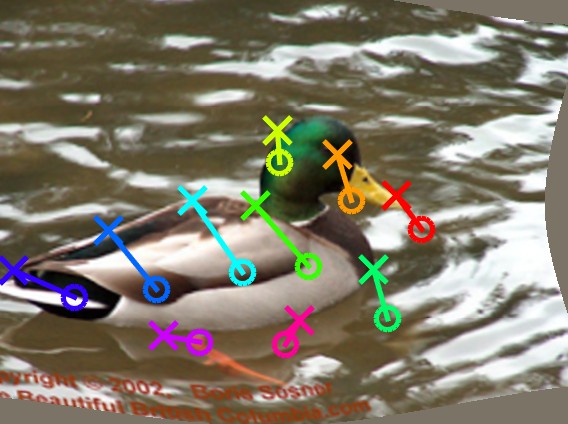}&
 \includegraphics[width=0.24\linewidth]{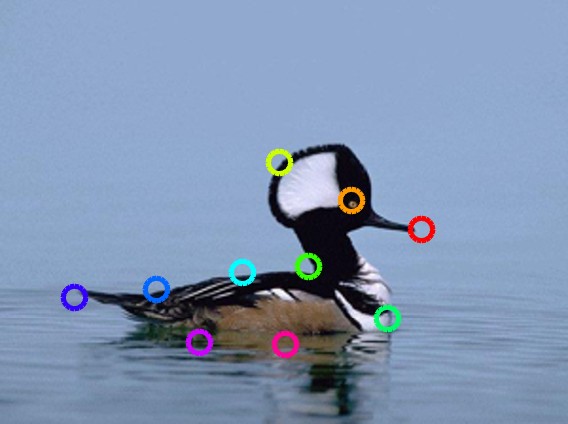}\\
 
 \includegraphics[width=0.24\linewidth]{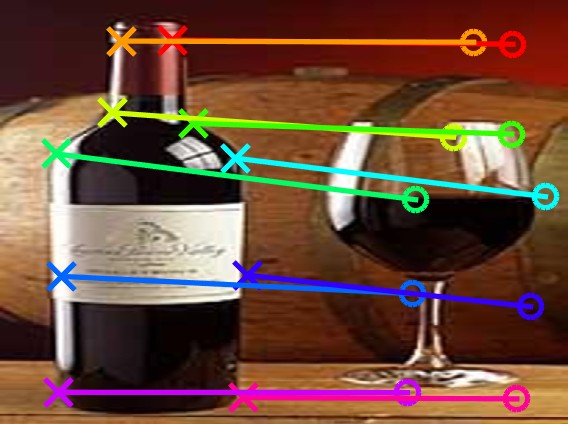}&
 \includegraphics[width=0.24\linewidth]{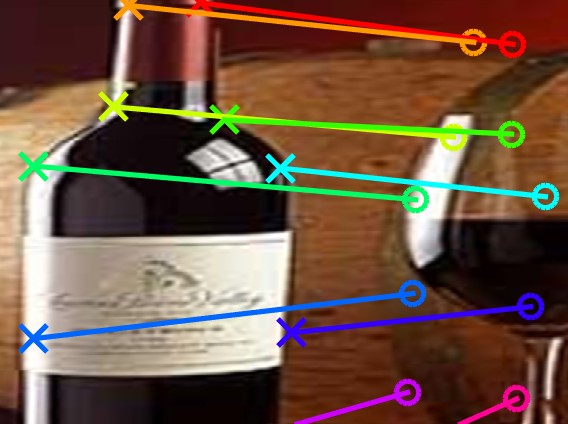}&
 \includegraphics[width=0.24\linewidth]{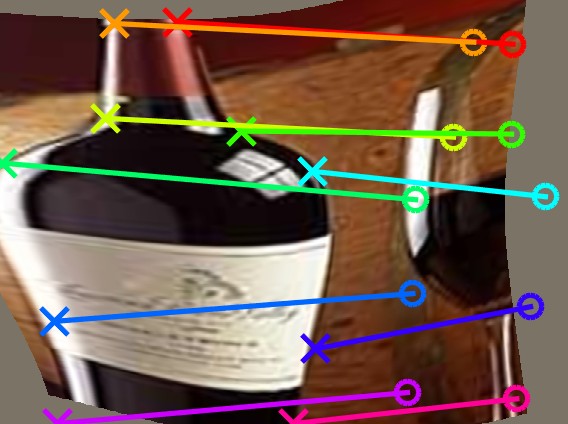}&
 \includegraphics[width=0.24\linewidth]{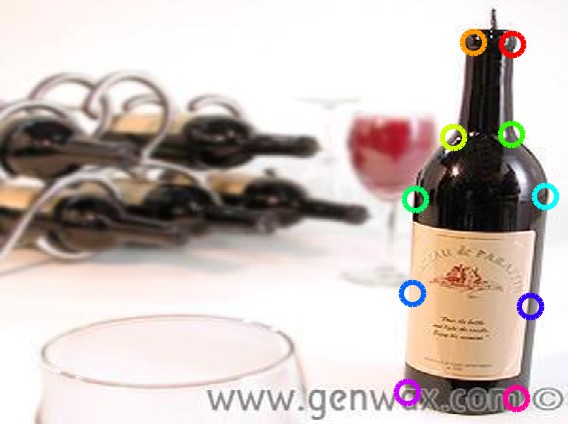}\\
 
 Image A&
 Aligned A (affine)&
 Aligned A (affine+TPS)&
 Image B
\end{tabular}
    \caption{{\bf Difficult examples from the Proposal Flow dataset.} Some examples with a combination of a significant change in viewpoint, appearance variations and/or background clutter are still challenging for our method, leading to only partial alignment (rows 1-3) or even a mis-alignment (row 4).}\label{fig:PF-difficult}
\end{figure*}

\vspace{5mm}
\section{Results on the Caltech-101 dataset \label{apx:caltech_results}}
In addition to the Proposal Flow dataset, we evaluate our method on the Caltech-101 dataset \cite{fei2006one} using the same procedure as in \cite{ham2016}. The results shown here were obtained using the same model trained from synthetically transformed StreetView images, which we used for evaluation on the Proposal Flow dataset. No further training was done to target this particular dataset.

As no keypoint annotations are provided for the Caltech-101 dataset, other metrics are needed to assess the matching accuracy. As segmentations masks are provided, we follow~\cite{kim2013deformable} and evaluate the following metrics: label transfer accuracy (LT-ACC), intersection-over-union (IoU), and localization error (LOC-ERR).

For each of the 101 categories, 15 image pairs were chosen randomly, resulting in 1515 evaluation pairs. These pairs match those used in \cite{ham2016}. As can be seen in Tab.\ \ref{tab:caltech}, our approach outperforms the state-of-the-art by a significant margin obtaining, for example, an IoU of 0.56 compared to the previous best result of 0.50. In Fig. \ref{fig:caltech}, we present a qualitative comparison of the results obtained by our method and other previous methods on images from this dataset.

\begin{table*}
\vspace*{4mm}
\centering
\begin{tabular}{lccc}
\hline
Methods & LT-ACC & IoU & LOC-ERR \\
\hline\hline
DeepFlow \cite{revaud2015deepmatching} & 0.74 & 0.40 & 0.34 \\
GMK \cite{Duchenne11} & 0.77 & 0.42 & 0.34 \\
SIFT Flow \cite{liu2011sift} & 0.75 & 0.48 & 0.32\\
DSP \cite{Kim13} & 0.77 & 0.47 & 0.35 \\
Proposal Flow (RP, LOM) \cite{ham2016} & 0.78 & 0.50 & 0.26 \\
Proposal Flow (SS, LOM) \cite{ham2016} & 0.78 & 0.50 & {\bf 0.25} \\
Ours (affine) & 0.79 & 0.51 & {\bf 0.25} \\
Ours (affine + thin-plate spline) & {\bf 0.82} & {\bf 0.56} & {\bf 0.25} \\
\hline \\
\end{tabular}
    \caption{{\bf Evaluation on the Caltech-101 dataset.}
Matching quality is measured in terms of LT-ACC, IoU and LOC-ERR.
The best two Proposal Flow methods (RP, LOM and SS, LOM) are included here. All numbers apart from ours are taken from \cite{ham2016}.
}
    \label{tab:caltech}
    \vspace{-0.3cm}
\end{table*}

\begin{figure*}
    \centering
\setlength\tabcolsep{1pt} 
\begin{tabular}{ccccccc}
 \includegraphics[width=0.135\linewidth]{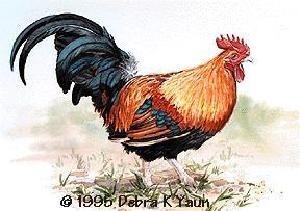}&
 \includegraphics[width=0.135\linewidth]{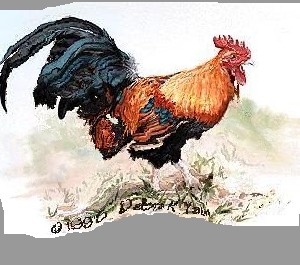}&
 \includegraphics[width=0.135\linewidth]{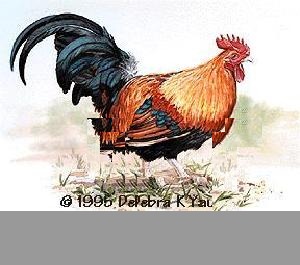}&
 \includegraphics[width=0.135\linewidth]{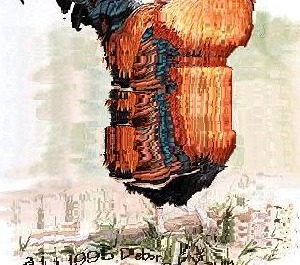}&
 \includegraphics[width=0.135\linewidth]{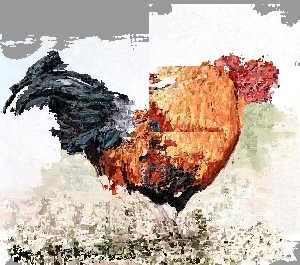}&
 \includegraphics[width=0.135\linewidth]{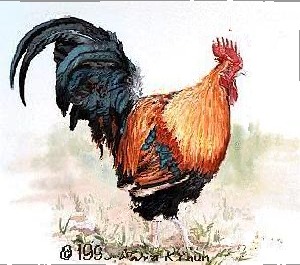}&
 \includegraphics[width=0.135\linewidth]{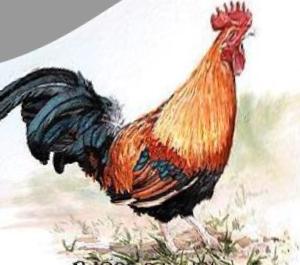}\\
  
 \includegraphics[width=0.135\linewidth]{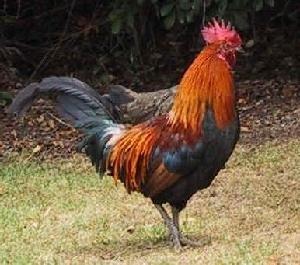}&
 \includegraphics[width=0.135\linewidth]{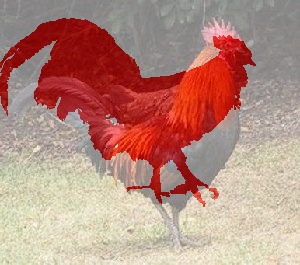}&
 \includegraphics[width=0.135\linewidth]{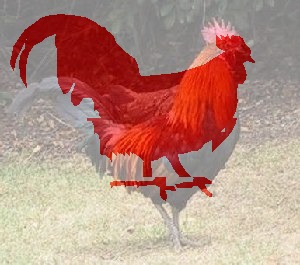}&
 \includegraphics[width=0.135\linewidth]{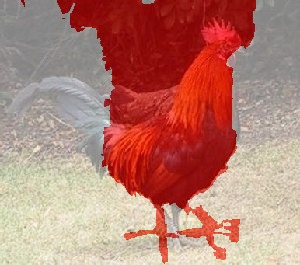}&
 \includegraphics[width=0.135\linewidth]{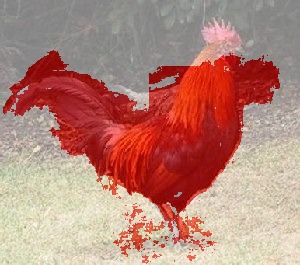}&
 \includegraphics[width=0.135\linewidth]{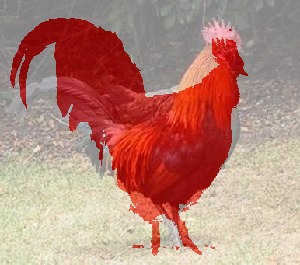}&
 \includegraphics[width=0.135\linewidth]{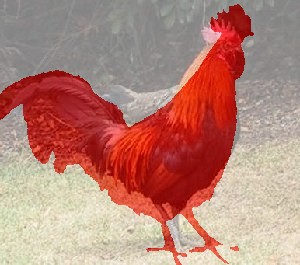} \\ \hdashline \vspace{-0.3cm} \\ 

 \includegraphics[width=0.135\linewidth]{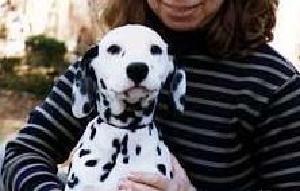}&
 \includegraphics[width=0.115\linewidth]{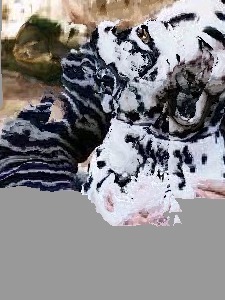}&
 \includegraphics[width=0.115\linewidth]{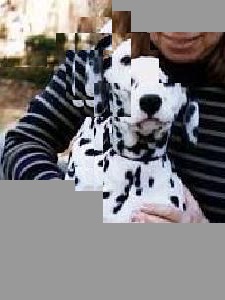}&
 \includegraphics[width=0.115\linewidth]{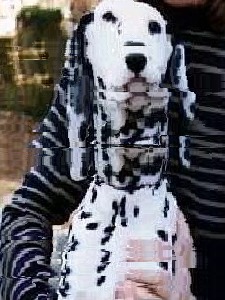}&
 \includegraphics[width=0.115\linewidth]{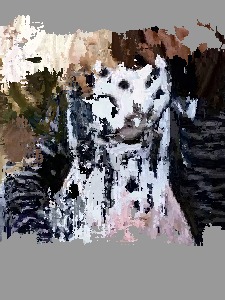}&
 \includegraphics[width=0.115\linewidth]{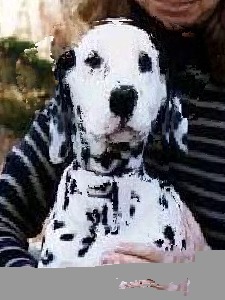}&
 \includegraphics[width=0.115\linewidth]{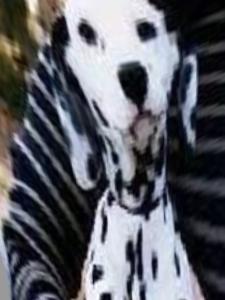}\\
  
 \includegraphics[width=0.115\linewidth]{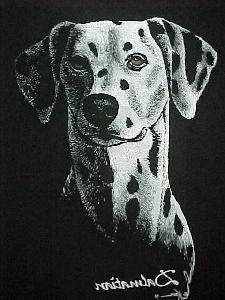}&
 \includegraphics[width=0.115\linewidth]{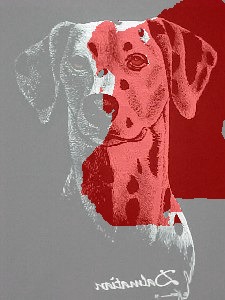}&
 \includegraphics[width=0.115\linewidth]{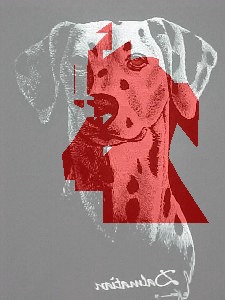}&
 \includegraphics[width=0.115\linewidth]{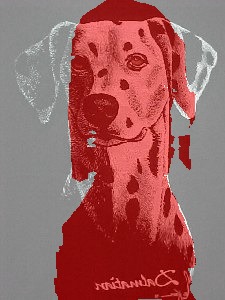}&
 \includegraphics[width=0.115\linewidth]{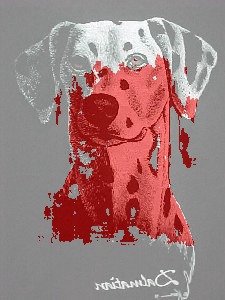}&
 \includegraphics[width=0.115\linewidth]{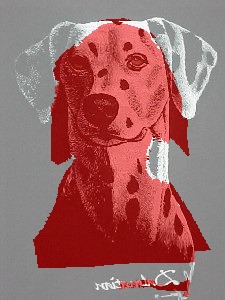}&
 \includegraphics[width=0.115\linewidth]{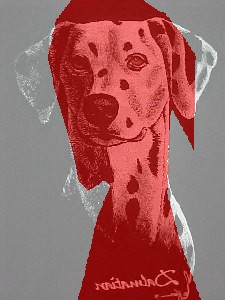}\\ \hdashline \vspace{-0.3cm} \\
 
 \includegraphics[width=0.135\linewidth]{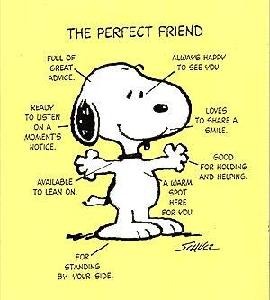}&
 \includegraphics[width=0.115\linewidth]{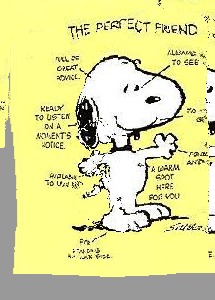}&
 \includegraphics[width=0.115\linewidth]{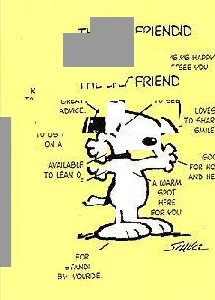}&
 \includegraphics[width=0.115\linewidth]{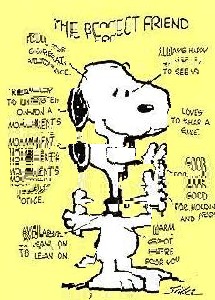}&
 \includegraphics[width=0.115\linewidth]{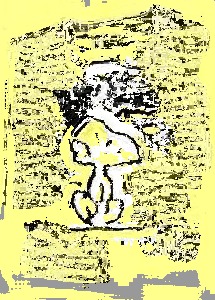}&
 \includegraphics[width=0.115\linewidth]{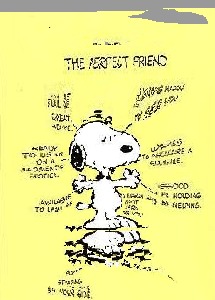}&
 \includegraphics[width=0.115\linewidth]{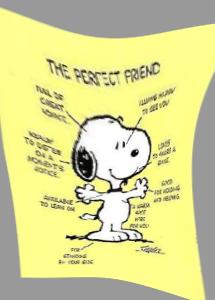}\\
  
 \includegraphics[width=0.115\linewidth]{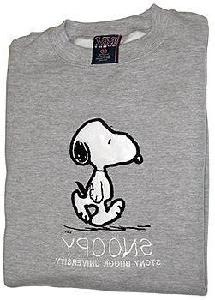}&
 \includegraphics[width=0.115\linewidth]{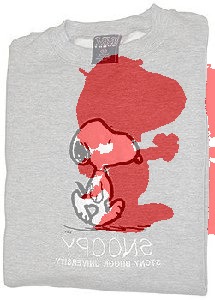}&
 \includegraphics[width=0.115\linewidth]{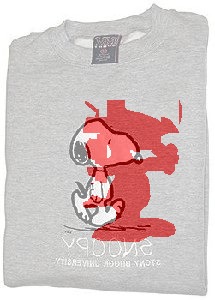}&
 \includegraphics[width=0.115\linewidth]{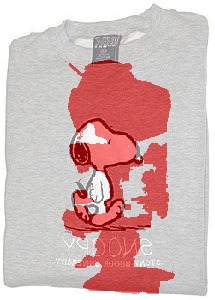}&
 \includegraphics[width=0.115\linewidth]{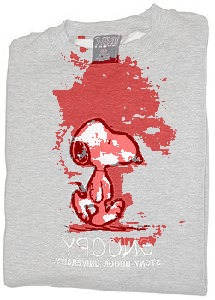}&
 \includegraphics[width=0.115\linewidth]{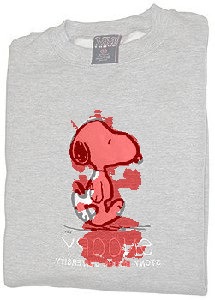}&
 \includegraphics[width=0.115\linewidth]{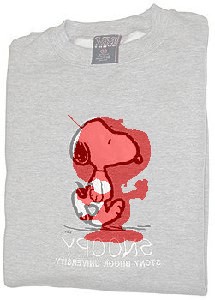}\\ \hdashline \vspace{-0.3cm}  \\
 
 \includegraphics[width=0.1\linewidth]{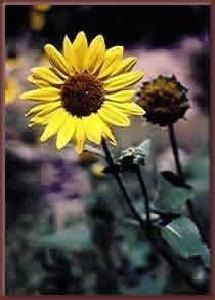}&
 \includegraphics[width=0.135\linewidth]{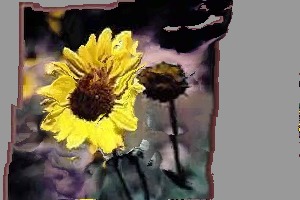}&
 \includegraphics[width=0.135\linewidth]{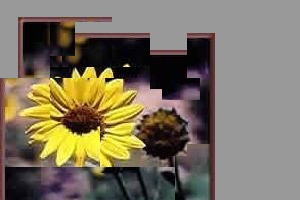}&
 \includegraphics[width=0.135\linewidth]{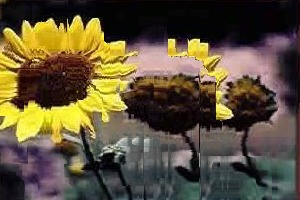}&
 \includegraphics[width=0.135\linewidth]{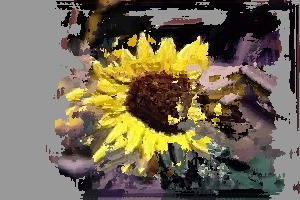}&
 \includegraphics[width=0.135\linewidth]{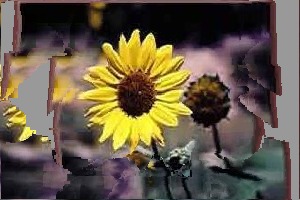}&
 \includegraphics[width=0.135\linewidth]{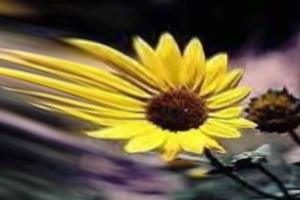}\\
  
 \includegraphics[width=0.135\linewidth]{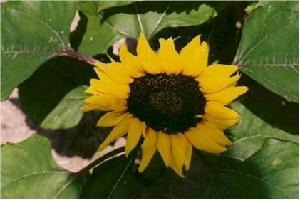}&
 \includegraphics[width=0.135\linewidth]{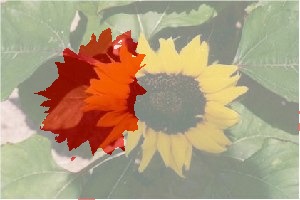}&
 \includegraphics[width=0.135\linewidth]{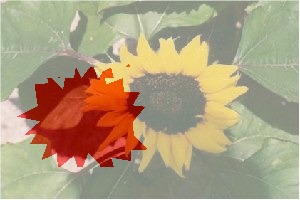}&
 \includegraphics[width=0.135\linewidth]{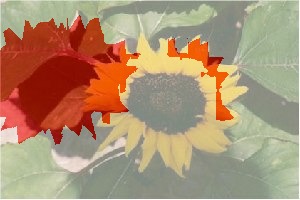}&
 \includegraphics[width=0.135\linewidth]{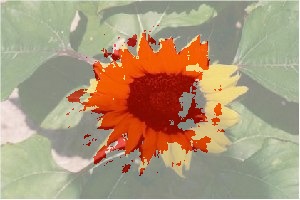}&
 \includegraphics[width=0.135\linewidth]{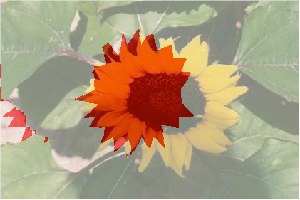}&
 \includegraphics[width=0.135\linewidth]{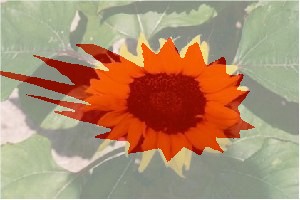}\\ 

{\small (a) Image pairs} & {\small (b) DeepFlow} & {\small (c) GMK} & {\small (d) SIFT Flow} & {\small (e) DSP} & {\small (f) Proposal Flow} & {\small (g) Our method} \\ 
\end{tabular}
    \caption{{\bf Qualitative examples from the Caltech-101 dataset.}
Each block of two rows corresponds to one example, where column (a) shows the original images -- image A in the first row and image B in the second row. The remaining columns of the first row show image A aligned to image B using various methods. The second row shows image B overlaid with the segmentation map transferred from image A.
}
\label{fig:caltech}
\end{figure*}

\section{Thin-plate spline transformation \label{apx:tps}}
The thin-plate spline (TPS) transformation \cite{bookstein1989principal} is a parametric model, which allows to perform 2D interpolation based on a set of known corresponding control points in the two images. In this work we use a fixed uniform $3\times 3$ grid of control points, which is defined over image B (as inverse sampling is used) and their corresponding points in image A. This is illustrated in Fig.\ \ref{fig:tps}. As the control points in image B are fixed for all image pairs, the TPS transformation is parametrized only by the control point positions in image A. Our TPS regression network estimates a 18-dimensional vector $\hat{\theta}_{TPS}$ composed of the 9 $x$-coordinates, followed by the 9 $y$-coordinates of these control points.

\begin{figure}[!h]
    \centering
    \begin{subfigure}[b]{0.3\textwidth}
        \includegraphics[width=\textwidth]{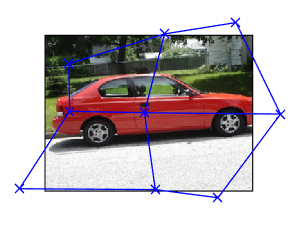}
        \caption{Control points over image A}
    \end{subfigure}
    ~ 
    \begin{subfigure}[b]{0.3\textwidth}
        \includegraphics[width=\textwidth]{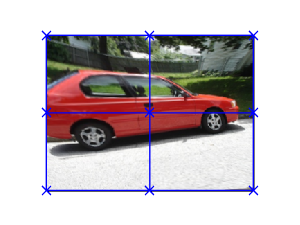}
        \caption{Control points over image B}
    \end{subfigure}
    \caption{Illustration of the $3\times 3$ TPS grid of control points used in our thin-plate spline transformation model.}\label{fig:tps}
\end{figure}

}{}

\end{document}